\documentclass{article} % For LaTeX2e
\usepackage[final]{colm2025_conference}

\usepackage{microtype}
\usepackage{hyperref}
\usepackage{url}
\usepackage{booktabs}
\usepackage{graphicx}
\usepackage{subfigure}

\usepackage{lineno}

\usepackage{amsmath}
\usepackage{amssymb}
\usepackage{mathtools}
\usepackage{amsthm}
\usepackage{wrapfig}

\theoremstyle{plain}

\theoremstyle{definition}

\theoremstyle{remark}

\definecolor{darkblue}{rgb}{0, 0, 0.5}
\hypersetup{colorlinks=true, citecolor=darkblue, linkcolor=darkblue, urlcolor=darkblue}

\title{Spike No More: \\ Stabilizing the Pre-training of Large Language Models}

\author{Sho Takase \& Shun Kiyono \\
SB Intuitions \\
\texttt{\{sho.takase,shun.kiyono\}@sbintuitions.co.jp} \\
\And
Sosuke Kobayashi \& Jun Suzuki \\
Tohoku University \\
\texttt{sosk@preferred.jp} \\
\texttt{jun.suzuki@tohoku.ac.jp} \\
}

\begin{document}

\ifcolmsubmission
\linenumbers
\fi

\maketitle

\begin{abstract}
Loss spikes often occur during pre-training of large language models.
The spikes degrade the performance of large language models and sometimes ruin the pre-training.
Since the pre-training needs a vast computational budget, we should avoid such spikes.
Based on the assumption that the loss spike is caused by the sudden growth of the gradient norm, we explore factors to keep the gradient norm small through an analysis of the spectral norms of the Jacobian matrices for the sub-layers.
Our findings suggest that stabilizing the pre-training process requires two conditions: small sub-layers and large shortcut.
We conduct various experiments to empirically verify our theoretical analyses.
Experimental results demonstrate that methods satisfying the conditions effectively prevent loss spikes during pre-training.\end{abstract}

\section{Introduction}
\label{sec:intro}

\begin{wrapfigure}[20]{r}[0pt]{6cm}
  \centering 
  \vskip -0.1in
    \includegraphics[width=6cm]{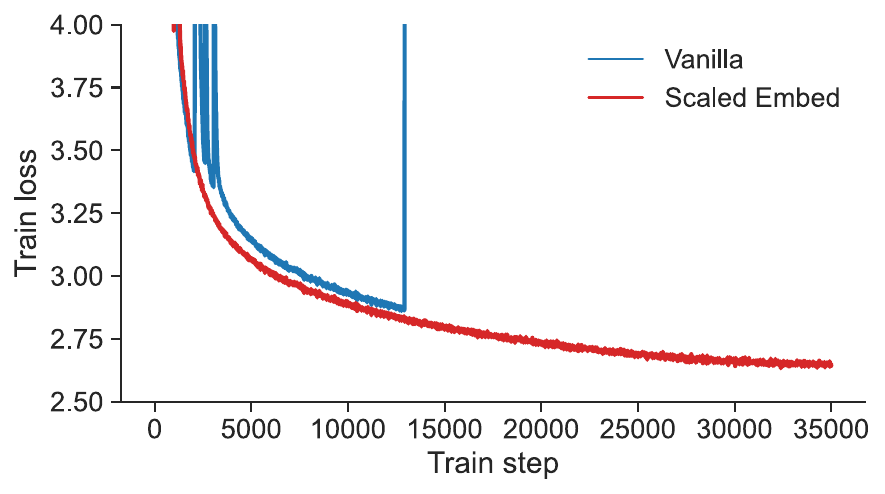}
   \caption{Training loss values of Transformers, whose dimensions and the number of layers are the same as the 1.7 billion parameters configuration in \citet{10.1145/3458817.3476209}. In Vanilla, some spikes occur at the beginning of the training, and its loss value exploded at about 13000 steps.}
   \label{fig:train_loss_explosion}
\end{wrapfigure}

Large language models (LLMs) have been fundamental assets for various applications~\citep{NEURIPS2020_1457c0d6,chowdhery2022palm,touvron2023llama}.
Increasing the number of parameters in (neural) language models and the number of training data usually leads to better LLMs~\citep{kaplan2020scaling}.
Consequently, pre-training requires a vast budget, and thus, minimizing the risk of failure of the pre-training is a paramount concern.

Despite their widespread use as the foundational architecture for LLMs, a comprehensive theoretical understanding of Transformers~\citep{NIPS2017_7181} has not yet been achieved. 
One of the crucial unresolved questions is the reason for the frequent occurrence of pre-training failures in Transformer-based LLMs due to spikes in loss values (loss spike) that can lead to catastrophic divergence~\citep{chowdhery2022palm} as illustrated in Vanilla in Figure \ref{fig:train_loss_explosion}.
While several empirical strategies have been proposed to mitigate this problem~\citep{chowdhery2022palm,le-scao-etal-2022-language,zeng2023glmb}, the absence of theoretical justification for these methods casts unclear on their generalizability to other situations, such as varying sizes of model parameters.

In this research, we provide theoretical analyses focusing on the loss spike problem during LLM pre-training.
We identify the upper bound of the gradient norms for the Transformer-based LLMs through analyses on the spectral norms of the Jacobian matrices for the sub-layers.
If the upper bound is large, the gradients may spike suddenly, and we assume that this phenomenon causes the loss spike.
Then, we indicate that the upper bound is large in the typical setting, such as the widely used implementation, Megatron-LM~\citep{shoeybi2020megatronlm}, and thus, the loss spike is likely to occur.
In addition, to make the upper bound sufficiently small, we introduce two conditions: (1) initializing the parameters of sub-layers with a small value and (2) making the standard deviation of each embedding close to $1$.
The former condition can be satisfied by the widely used initialization method for LLMs~\citep{shoeybi2020megatronlm,le-scao-etal-2022-language,10.5555/3618408.3618510}.
On the other hand, the latter condition was satisfied in the original Transformer by scaling embeddings~\citep{NIPS2017_7181}, but such scaling is missing from recent implementations.
To sum up, through theoretical analyses, we re-evaluate several previous techniques in terms of the stabilization of LLM pre-training.

Building on our theoretical analysis, we further substantiate our claims through a series of empirical experiments, which provides a clear distinction between effective and ineffective methods over different training scenarios.
Our results demonstrate that methods satisfying the conditions avoid the occurrence of loss and gradient spikes.
In contrast, methods that fail to meet these conditions remain susceptible to gradient spikes, even when previously recommended as empirical solutions of the loss spike problem.
Furthermore, we demonstrate that a method satisfying the conditions enables LLMs to be pre-trained with a comparatively larger learning rate, leading to superior performance outcomes.

\section{Preliminary}
\label{sec:preliminary}

\subsection{Pre-LN Transformer}
\label{sec:preln}
This paper mainly focuses on the neural architecture used in the GPT series~\citep{radford2018improving,Radford2019LanguageMA,NEURIPS2020_1457c0d6}.
They use the Pre-LN Transformer~\citep{DBLP:conf/icml/XiongYHZZXZLWL20}, which is the de facto standard architecture in recent implementations of Transformers because the training with the architecture is more stable than the original Transformer architecture when we stack many layers~\citep{DBLP:conf/icml/XiongYHZZXZLWL20,liu-etal-2020-understanding,takase-etal-2023-b2t}.
Let $x \in \mathbb{R}^{d}$ be an input of a layer of the Transformer, where $d$ denotes the dimension of the layer.
The layer outputs $y$ with the following equations:
\begin{align}
y &= x' + \mathrm{FFN}(\mathrm{LN}(x')) , \label{eq:preln_ffn} \\
x' &= x + \mathrm{Attn}(\mathrm{LN}(x)) , \label{eq:preln_attn}
\end{align}
where $\mathrm{LN}$ is the layer normalization function\footnote{We discuss the difference from the architecture using Root Mean Square layer normalization (RMSNorm)~\citep{NEURIPS2019_1e8a1942} instead of LN in Appendix \ref{sec:rmsnorm}, and the original Transformer architecture, i.e., Post-LN Transformer in Appendix \ref{sec:post-ln}.}.
We call the first terms in Equations (\ref{eq:preln_ffn}) and (\ref{eq:preln_attn}), i.e., $x$ and $x'$, \textbf{shortcut}.
In addition, the feed-forward network ($\mathrm{FFN}$) and multi-head self-attention ($\mathrm{Attn}$) are defined as follows\footnote{To simplify equations, we omit bias terms.}:
\begin{align}
\mathrm{FFN}(x) &= W_2(\mathcal{F}(W_1 \ x)), \label{eq:ffn} \\
\mathrm{Attn}(x) &= W_O(\mathrm{concat}(\mathrm{head}_1(x), ..., \mathrm{head}_h(x))), \label{eq:attn} \\ 
%\nonumber 
\mathrm{head}_i(x) &= \mathrm{softmax}\left(\frac{(W_{Qi} \ x)^\mathrm{T} (W_{Ki} \ X)}{\sqrt{d_{\mathrm{head}}}}\right) (W_{Vi} \ X)^\mathrm{T} , \label{eq:attn_head}
\end{align}
where $\mathcal{F}$ is an activation function, $\mathrm{concat}$ concatenates input vectors, $\mathrm{softmax}$ applies the softmax function to an input vector, and $W_1 \in \mathbb{R}^{d_{\mathrm{ffn}} \times d}$, $W_2 \in \mathbb{R}^{d \times d_{\mathrm{ffn}}}$, $W_{Qi} \in \mathbb{R}^{d_{\mathrm{head}} \times d}$, $W_{Ki} \in \mathbb{R}^{d_{\mathrm{head}} \times d}$, $W_{Vi} \in \mathbb{R}^{d_{\mathrm{head}} \times d}$, and $W_{O} \in \mathbb{R}^{d \times d}$ are parameter matrices, and $d_{\mathrm{ffn}}$ and $d_{\mathrm{head}}$ are the internal dimensions of FFN and multi-head self-attention sub-layers, respectively.
In addition, we pack the sequence of input vectors into a matrix as $X \in \mathbb{R}^{d \times L}$, where $L$ is the input sequence length, to compute the self-attention.

\subsection{Gradients of Pre-LN Transformers}
\label{sec:gradients_llm}
Let $\mathcal{L}$ be the loss function of the $N$ layered Pre-LN Transformer and $J_n$ be the Jacobian matrix of the $n$-th layer.
We can calculate the gradient of $\mathcal{L}$ using the relations in Equations (\ref{eq:preln_ffn}) and (\ref{eq:preln_attn}) as:
\begin{align}
&\frac{\partial \mathcal{L}}{\partial x_1} %&= \frac{\partial \mathcal{L}}{\partial y_N}\ J_{N-1} ... \ J_{1} \\
= \frac{\partial \mathcal{L}}{\partial y_N} \prod_{n=1}^{N-1} J_{n}
= \frac{\partial \mathcal{L}}{\partial y_N} \prod_{n=1}^{N-1} \left(
\frac{\partial y_n}{\partial x'_n}\frac{\partial x'_n}{\partial x_n}
\right)
,\quad 
\mbox{where}
\quad
J_n = \frac{\partial y_n}{\partial x_n} = \frac{\partial y_n}{\partial x'_n}\frac{\partial x'_n}{\partial x_n}
.
\label{eq:L_grad}
\end{align}

Using the submultiplicativity of the spectral norm, i.e., $\|AB\|_2 \leq \|A\|_2\|B\|_2$, and Equation (\ref{eq:L_grad}), we can derive an upper bound of the norm of the gradient of $\mathcal{L}$ as:
\begin{align}
\bigg\| \frac{\partial \mathcal{L}}{\partial x_1} \bigg\|_2
= \bigg\| \frac{\partial \mathcal{L}}{\partial y_N} \prod_{n=1}^{N-1} \frac{\partial y_n}{\partial x'_n}  \frac{\partial x'_n}{\partial x_n}\bigg\|_2
 \leq \bigg\| \frac{\partial \mathcal{L}}{\partial y_N} \bigg\|_2 \prod_{n=1}^{N-1} \bigg\|\frac{\partial y_n}{\partial x'_n} \bigg\|_2 \bigg\| \frac{\partial x'_n}{\partial x_n}\bigg\|_2
.
\label{eq:upper_bound_transformer}
\end{align}
Thus, we can estimate the upper bound of the gradient norm of $\mathcal{L}$ by analyzing the spectral norms of the Jacobian matrices for the FFN layer and the self-attention layer, namely, $\|\frac{\partial y_n}{\partial x'_n} \|_2$ and $\| \frac{\partial x'_n}{\partial x_n}\|_2$.

\subsection{Motivation to Suppress the Upper Bound}\label{sec:motivation}
In our preliminary experiments, when the gradient norms grow suddenly during LLM pre-training, we observe that the loss spike problem is likely to occur.
Thus, we assume that we can prevent the loss spike problem by maintaining the gradient norm small.
To prevent the growth of the gradient norm, we explore the way to suppress the upper bound described by Equation (\ref{eq:upper_bound_transformer}).
To suppress the upper bound, we analyze the Jacobian matrices to find a factor to control the upper bound in the following sections, and then, provide two conditions: \textbf{small sub-layers} and \textbf{large shortcut}.
We verify our assumption and theoretical analyses through experiments on LLM pre-training.

\section{Analyses on Gradients of Sub-Layers}
\label{sec:theory}
For the theoretical analyses in this section, we employ the following assumption:
\paragraph{Assumption 1.}
Let $x$ and $x'$ be the input and intermediate vectors of each layer. Moreover, let $W_{*}$ denote the model parameter matrix in each layer. 
We assume that $x$, $x'$, and $W_{*}$ for all layers follow a normal distribution with a mean of 0, i.e., $\mu = 0$.

This assumption is valid when we initialize parameters with the normal distribution, the number of heads in Equation (\ref{eq:attn}) is $1$, and $\mathcal{F}$ is an identity function. Empirically, the outputs of each sub-layer are close to the normal distribution as illustrated in Appendix \ref{sec:sublayer_distribution}.

\subsection{Jacobian Matrix of FFN}
\label{sec:analysis_on_jacobian_ffn}
Based on Equation (\ref{eq:preln_ffn}), $\|\frac{\partial y_n}{\partial x'_n} \|_2$ in Equation (\ref{eq:upper_bound_transformer}) can be rewritten as:
\begin{align}
%\nonumber 
\bigg\|\frac{\partial y}{\partial x'} \bigg\|_2 &= \bigg\|\frac{\partial (x' + \mathrm{FFN}(\mathrm{LN}(x')))}{\partial x'} \bigg\|_2
= \bigg\| I + \frac{\partial (\mathrm{FFN}(\mathrm{LN}(x')))}{\partial x'} \bigg\|_2 
\end{align}
We can then derive an upper bound of $\|\frac{\partial y_n}{\partial x'_n} \|_2$ by applying the subadditivity, i.e., $\|A + B\|_2 \leq \| A \|_2 + \| B \|_2$, and submultiplicativity properties of the spectral norm as follows:
\begin{align}
%\nonumber 
\bigg\|\frac{\partial y}{\partial x'} \bigg\|_2
& \leq 1 + \bigg\| \frac{\partial \mathrm{FFN}(\mathrm{LN}(x'))}{\partial \mathrm{LN}(x')} \bigg\|_2 \bigg\|\frac{\partial \mathrm{LN}(x')}{\partial x'}\bigg\|_2
.
\label{eq:ffn_detail_upper_bound}
\end{align}
The right-hand side of this inequality indicates that we can estimate the upper bound of $\|\frac{\partial y}{\partial x'} \|_2$ by separately computing the spectral norms of Jacobian matrices for $\mathrm{FFN}$ and $\mathrm{LN}$.

Regarding the $\mathrm{FFN}$ part,
we assume that the activation function $\mathcal{F}$ is an identity function\footnote{Appendix \ref{sec:relu_in_ffn} discusses the case where we use the $\mathrm{ReLU}$, $\mathrm{SiLU}$, and $\mathrm{SwiGLU}$ as the activation functions, which leads to the same conclusion.}
to simplify the discussion.
Under this assumption, the following equation holds:
\begin{align}
 \bigg\| \frac{\partial \mathrm{FFN}(\mathrm{LN}(x'))}{\partial \mathrm{LN}(x')}\bigg\|_2 = \|W_2 W_1 \|_2 
.
\label{eq:ffn_spectral_norm}
\end{align}
Therefore, we can straightforwardly derive the relation $\| W_2 W_1 \|_2 \leq \|W_1\|_2 \|W_2\|_2$ from the submultiplicativity of the spectral norm.
Furthermore, let $\sigma_1$ and $\sigma_2$ be the standard deviations of $W_1$ and $W_2$, respectively.
From Assumption 1, the spectral norms of $W_1$ and $W_2$ are obtained by their standard deviations and dimensions~\citep{Vershynin_2018}, i.e., $\|W_1\|_2 \approx \sigma_1 (\sqrt{d} + \sqrt{d_{\mathrm{ffn}}})$ and $\|W_2\|_2 \approx \sigma_2 (\sqrt{d} + \sqrt{d_{\mathrm{ffn}}})$.
Finally, we can express an upper bound of the spectral norms of the Jacobian matrices for $\mathrm{FFN}$ as the following inequality:
\begin{align}
%\nonumber 
\bigg\| \frac{\partial \mathrm{FFN}(\mathrm{LN}(x'))}{\partial \mathrm{LN}(x')}\bigg\|_2 
 \leq 
\sigma_1 \sigma_2 (\sqrt{d} + \sqrt{d_{\mathrm{ffn}}})^2 
,
\label{eq:ffn_spectral_norm_final}
\end{align}
where the right-hand side has the relation $\sigma_1 \sigma_2 (\sqrt{d} + \sqrt{d_{\mathrm{ffn}}})^2  \approx \|W_1\|_2 \| W_2 \|_2$.

Next, regarding the $\mathrm{LN}$ part, the Jacobian matrix of LN can be written as:
\begin{align}
\frac{\partial \mathrm{LN}(x')}{\partial x'} 
= \frac{\sqrt{d}}{\| x' \|_2}\bigg(I - \frac{x'x'^\top}{\| x' \|_2^2} \bigg) 
= \frac{\sqrt{d}}{\sigma_{x'}\sqrt{d}}\bigg(I - \frac{x' x'^\top}{\sigma_{x'}^2 d} \bigg)
= \frac{1}{\sigma_{x'}}\bigg(I - \frac{zz^\top}{d} \bigg)
.
\label{eq:ln_spectral_norm}
\end{align}
The leftmost equation appears in the proof by \citet{DBLP:conf/icml/XiongYHZZXZLWL20}. 
The second equation uses $\| x' \|_2 = \sigma_{x'} \sqrt{d}$, which can be obtained based on Assumption 1.
The last equation is derived from the well-known formula of $z=(x'-\mu_{x'})/\sigma_{x'}$, which converts a normal distribution, $x'$, to the standard normal distribution $z$, where $\mu_{x'}=0$ in Assumption 1.

We consider the variance ($\mathrm{var}$) of each element in the matrix $zz^\top$.
Since $z_{i}z_{i}$ follows $\mathcal{X}^2$ with $1$ degree of freedom, and $z_{i}z_{j} (i \neq j)$ is the multiplication of two independent values following the standard normal distribution, the variances are as follows:
\begin{align}
\mathrm{var}(z_{i}z_{j}) = 
\begin{cases}
1 \ \  \text{if $i \neq j$} \\
2 \ \ \text{otherwise}
\end{cases}
.
\label{eq:variance_LN}
\end{align}
Equation (\ref{eq:variance_LN}) indicates that $\frac{zz^\top}{d} \approx 0$ in LLMs due to $d \gg 1$.
Therefore, the spectral norm of the Jacobian matrix of LN can be written as: 
\begin{align}
\bigg\|\frac{\partial \mathrm{LN}(x')}{\partial x'}\bigg\| = \frac{1}{\sigma_{x'}} 
,\quad\mbox{where}\quad
\frac{\partial \mathrm{LN}(x')}{\partial x'} = \frac{1}{\sigma_{x'}}I
.
\label{eq:ln_jacobian}
\end{align}

Finally, Equation (\ref{eq:ffn_detail_upper_bound}) can be rewritten by substituting Equations (\ref{eq:ffn_spectral_norm_final}) and (\ref{eq:ln_jacobian}) as:
\begin{align}
\bigg\|\frac{\partial y}{\partial x'}\bigg\|_2 &\leq 
1 + \frac{\sigma_1 \sigma_2}{\sigma_{x'}} C_{\mathrm{ffn}} , \label{eq:final_upper_bound_ffn}
\end{align}
where $C_{\mathrm{ffn}} = (\sqrt{d} + \sqrt{d_{\mathrm{ffn}}})^2$ for the simplification.

According to the discussion in Section~\ref{sec:motivation} and Equation (\ref{eq:final_upper_bound_ffn}), 
the standard deviations, $\sigma_1$ and $\sigma_2$, of $W_1$ and $W_2$, respectively, should be sufficiently small, and the standard deviation, $\sigma_{x'}$, of the shortcut, $x'$, should satisfy $\sigma_1\sigma_2 \ll \sigma_{x'}$ in order to keep the upper bound small.

\subsection{Jacobian Matrix of Self-Attention}
\label{sec:analysis_on_jacobian_attn}
Similar to $\mathrm{FFN}$, we can rewrite $\|\frac{\partial x'}{\partial x} \|_2$ in Equation (\ref{eq:upper_bound_transformer}) by using Equation (\ref{eq:preln_attn}) as:
\begin{align}
%\nonumber 
\bigg\|\frac{\partial x'}{\partial x} \bigg\|_2 &= \bigg\|\frac{\partial (x + \mathrm{Attn}(\mathrm{LN}(x)))}{\partial x} \bigg\|_2
= \bigg\| I + \frac{\partial (\mathrm{Attn}(\mathrm{LN}(x)))}{\partial x} \bigg\|_2 
.
\end{align}
We can then derive an upper bound of $\|\frac{\partial x'}{\partial x} \|_2$ by applying the subadditivity and submultiplicativity of the spectral norm, namely:
\begin{align}
%\nonumber
 \bigg\|\frac{\partial x'}{\partial x} \bigg\|_2 
& \leq 1 + \bigg\| \frac{\partial \mathrm{Attn}(\mathrm{LN}(x))}{\partial \mathrm{LN}(x)} \bigg\|_2 \bigg\|\frac{\partial \mathrm{LN}(x)}{\partial x}\bigg\|_2 
.
\label{eq:attn_detail_upper_bound}
\end{align}
Therefore, to estimate the upper bound of $\|\frac{\partial x'}{\partial x} \|_2$, we compute the spectral norms of the Jacobian matrices for $\mathrm{Attn}$ and $\mathrm{LN}$.

Let $Z(\cdot) = \mathrm{concat}(\mathrm{head}_1(\cdot), ..., \mathrm{head}_h(\cdot)))$ and let $J^{Z}$ be the Jacobian of the $Z(\cdot)$\footnote{We discuss the detail of $J^{Z}$ in Appendix \ref{sec:detail_selfattn_jacobian}.}, we can rewrite the spectral norm of the Jacobian matrix of Attn as:
\begin{align}
%\nonumber 
\bigg\|\frac{\partial \mathrm{Attn}(\mathrm{LN}(x))}{\partial \mathrm{LN}(x)}\bigg\|_2 
= \bigg\| \frac{\partial W_O Z(\mathrm{LN}(x))}{\partial Z(\mathrm{LN}(x))}\frac{\partial Z(\mathrm{LN}(x))}{\partial \mathrm{LN}(x)}\bigg\|_2
= \| W_O J^{Z}\|_2 
.
\label{eq:multihead_attn_upper_bound}
\end{align}
Therefore, we can straightforwardly derive the relation $\| W_O J^{Z}\|_2 \leq \| W_O\|_2 \|J^{Z}\|_2$ from the submultiplicativity of the spectral norm.

Let $\sigma_O$ be the standard deviation of $W_O$.
The relation $\|W_O\|_2 \approx \sigma_O(2\sqrt{d})$ is derived from Assumption 1.
We assign this value to Equation (\ref{eq:multihead_attn_upper_bound}) and obtain the following inequality:
\begin{align}
%\nonumber 
\bigg\|\frac{\partial \mathrm{Attn}(\mathrm{LN}(x))}{\partial \mathrm{LN}(x)}\bigg\|_2 
\leq 
\sigma_O(2\sqrt{d})\|J^{Z}\|_2 .
\label{eq:tmp_attn_spectral_norm}
\end{align}

Therefore, we can rewrite Equation (\ref{eq:attn_detail_upper_bound}) by substituting Equations (\ref{eq:ln_jacobian}) and (\ref{eq:tmp_attn_spectral_norm}) as follows:
\begin{align}
\bigg\|\frac{\partial x'}{\partial x}\bigg\|_2 &\leq %1 + \frac{\sigma_O}{\sigma_{x}}(2\sqrt{d})\|J^{Z}\|_2 \\
%&= 
1 + \frac{\sigma_O}{\sigma_{x}} C_{\mathrm{Attn}}
,
\label{eq:final_upper_bound_attn}
\end{align}
where $C_{\mathrm{Attn}} = (2\sqrt{d})\|J^{Z}\|_2$ for the simplification.

Thus, similar to the discussion at the end of Section \ref{sec:analysis_on_jacobian_ffn}, the standard deviation, $\sigma_O$, of $W_O$ should be small and the standard deviation, $\sigma_{x}$, of the shortcut, $x$, should satisfy $\sigma_O \ll \sigma_{x}$ in order to keep the upper bound small.

\section{Conditions to Avoid Spikes}
\label{sec:proposed_method}
Based on the discussions in Section \ref{sec:theory}, we have to pay attention to values of $\sigma_1$, $\sigma_2$, $\sigma_O$, and the standard deviation of the shortcut to stabilize the pre-training of LLMs.
To make $\sigma_1$, $\sigma_2$, and $\sigma_O$ small, we have to initialize the corresponding parameters with a small value.
Let us consider the actual settings in detail.
The widely used initialization method for LLMs~\citep{shoeybi2020megatronlm,le-scao-etal-2022-language,10.5555/3618408.3618510}, initializes all parameters with a normal distribution $\mathcal{N}(0, \sigma^2)$ where $\sigma = \sqrt{\frac{2}{5d}}$~\citep{nguyen-salazar-2019-transformers}, and then scales $W_2$ and $W_O$ to small values based on the number of layers: $\sqrt{\frac{1}{2N}}$ where $N$ is the number of layers\footnote{\citet{10.5555/3618408.3618510} also scaled $W_2$ and $W_O$ to small values in the initialization, but they used the strategy introduced by \citet{gpt-j} instead of scaling with $\sqrt{\frac{1}{2N}}$. However, its property is the same essentially because they initialize $W_2$ and $W_O$ with $\sigma = \frac{2}{N\sqrt{d}}$ which becomes small based on the number of layers.}.
In this situation, $\sigma_1$, $\sigma_2$, and $\sigma_O$ are sufficiently small values.

However, in this situation, the standard deviation of the shortcut is also too small.
For example, at shallow layers, the standard deviation is close to $\sqrt{\frac{2}{5d}}$ because the embedding matrix is also initialized by $\mathcal{N}(0, \sigma^2)$ where $\sigma = \sqrt{\frac{2}{5d}}$.
Therefore, to increase the standard deviation of the shortcut, we make the standard deviation of each embedding close to $1$\footnote{Based on Equations (\ref{eq:final_upper_bound_ffn}) and (\ref{eq:final_upper_bound_attn}), the upper bound becomes small as the standard deviation of the shortcut increases. However, a too large value degrades the performance empirically as described in Appendix \ref{sec:scaling_larger_value}.}.
To achieve this, we introduce two kinds of modification: ``Scaled Embed'' and ``Embed LN''\footnote{We can satisfy the condition by initializing embeddings with the normal distribution $\mathcal{N}(0, \sigma^2)$ where $\sigma = 1$, but we do not adopt this strategy in this study because we use the same initialization method in our experiments.}.
The Scaled Embed scales embeddings with an appropriate value.
For example, we multiply embeddings by $\sqrt{d}$, which was used in the original Transformer paper~\citep{NIPS2017_7181}\footnote{Although the original Transformer paper introduced this operation, recent implementations ignore this. Therefore, LLMs trained with recent implementations do not satisfy large shortcut.}, and then the standard deviations of embeddings become $\sqrt{\frac{2}{5}}$.
The Embed LN applies the LN to embeddings.
In fact, \citet{le-scao-etal-2022-language} reported that the Embed LN strategy prevents loss spikes empirically.
These two methods are presented as verification examples rather than proposed methods, and alternative approaches could be employed if the conditions are met.

\begin{wrapfigure}[18]{r}[0pt]{7cm}
  \vskip -0.1in
  \centering 
    \includegraphics[width=7cm]{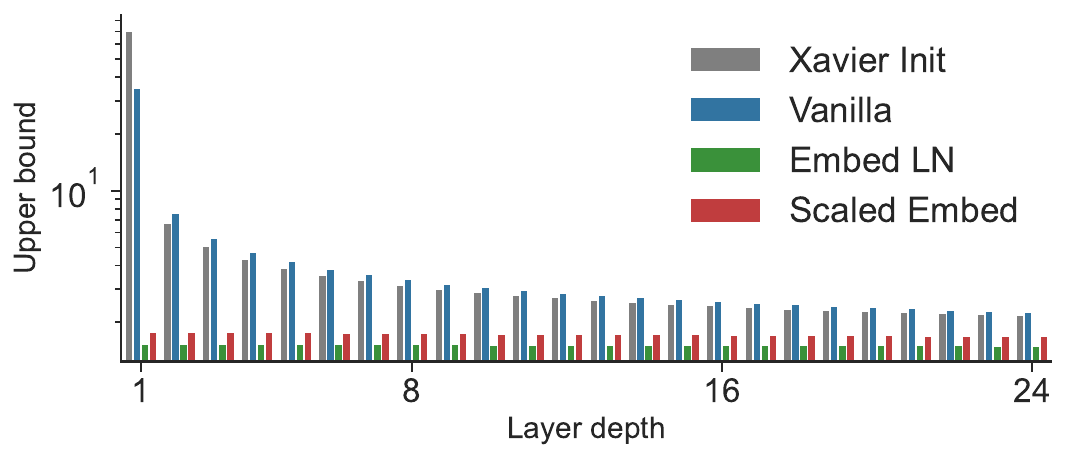}
   \caption{The actual upper bound described in Equation (\ref{eq:final_upper_bound_ffn}) for each Transformer layer at the beginning of the LLM pre-training. Because it is difficult to estimate the strict values for $\sigma_{x}$ at all layers, we obtain the empirical values by using some inputs, and assign them to Equation (\ref{eq:final_upper_bound_ffn}).}
   \label{fig:bound_each_layer}
\end{wrapfigure}

To demonstrate the actual values of the upper bound described in Equation (\ref{eq:final_upper_bound_ffn}), we take the model with 1.7 billion parameters as an example.
In addition to the widely used initialization for LLMs (Vanilla) and the above two modifications: Scaled Embed and Embed LN, we compare Xavier Init, which initializes all parameters with the Xavier initialization~\citep{pmlr-v9-glorot10a}, as the situation where we do not scale $W_2$ and $W_O$ based on the number of layers.
Figure \ref{fig:bound_each_layer} shows the values of Equation (\ref{eq:final_upper_bound_ffn}) for each layer at the beginning of the pre-training.
This figure indicates that the methods without suppressing the upper bound, i.e., Xavier Init and Vanilla, rapidly increase the values especially in shallow layers.
In contrast, Scaled Embed and Embed LN keep small values.
In summary, to make the upper bound of the gradient norms small for the stabilization of the LLM pre-training, we have to satisfy two conditions: (1) \textbf{small sub-layers}; initializing the parameters of sub-layers with a small value and (2) \textbf{large shortcut}; making the standard deviation of each embedding close to $1$.

\section{Experiments}
\label{sec:main_exp}
We verify the empirical effectiveness of our theoretical analyses.
In detail, we demonstrate that controlling the upper bound of the gradient norms also prevents loss and gradient spikes.
To assess efficacy in the real situation, we focus on the methods initialized with the widely used method~\citep{shoeybi2020megatronlm,le-scao-etal-2022-language} in main experiments\footnote{The Xavier initialization, which does not satisfy the small sub-layers condition, is not widely used for LLMs in recent years. Appendix \ref{sec:xavier_init} shows that the performance of Xavier initialization is worse and fails to avoid spikes.}.

\subsection{Datasets}
We used C4~\citep{JMLR:v21:20-074} that consists of clean English texts extracted from Common Crawl as our LLM pre-training corpus.
We also used the separated part of C4 as our validation data.
We used GPT-2 vocabulary~\citep{Radford2019LanguageMA} that contains Byte Pair Encoding (BPE) subword units~\citep{sennrich-etal-2016-neural} as our vocabulary.
To evaluate each method, we computed perplexity on WikiText~\citep{DBLP:journals/corr/MerityXBS16} and LAMBADA~\citep{paperno-etal-2016-lambada} datasets.

\subsection{Model Configurations}
As described in Section \ref{sec:preliminary}, we used the Pre-LN Transformer architecture.
We conduct experiments with two parameter sizes: 350 million (350M) and 1.7 billion (1.7B).
We set the learning rate ($\mathrm{lr}$) $5.0 \times 10^{-4}$.
Appendix \ref{sec:hyperparams} describes more details on the experimental configuration.
We compared the following methods.
We put $\checkmark$ before the method name if the method satisfies both conditions to suppress the upper bound.

\paragraph{Vanilla}
This is the standard configuration for LLM pre-training.
Since this configuration does not suppress the upper bound of the gradient norms, the loss spike is likely to occur.

\paragraph{Embed Detach}
\citet{zeng2023glmb} used the shrink embedding gradient technique~\citep{NEURIPS2021_a4d92e2c} to stabilize their LLM pre-training.
This method shrinks gradients on the embedding layer by detaching a part of embeddings from the computational graph as follows:
\begin{align}
\mathrm{Embed} \gets \gamma \mathrm{Embed} + (1 - \gamma) \mathrm{Detach}(\mathrm{Embed}),
\end{align}
where $\gamma$ is a hyper-parameter and $\mathrm{Detach}$ detaches an input from the computational graph.
We assign $0.1$ to $\gamma$ as in \citet{zeng2023glmb}.
\citet{zeng2023glmb} indicated that this method empirically prevents the loss spike.
However, this method does not satisfy the condition on large shortcut, and thus, we show that this method does not completely solve the loss spike.

\paragraph{$\checkmark$Embed LN}
\citet{dettmers2022bit} and \citet{le-scao-etal-2022-language} reported that applying the LN to the embedding layer stabilizes their LLM pre-training.
As described in Section \ref{sec:proposed_method}, this method satisfies the conditions to control the upper bound of the gradient norms.

\paragraph{$\checkmark$Scaled Embed}
This method multiplies embeddings by $\sqrt{d}$.
As described in Section \ref{sec:proposed_method}, this method satisfies the requirements to control the upper bound of the gradient norms.

\subsection{Main Results}
\label{sec:main_results}
\begin{figure}[!t]
\centering
    \subfigure[350M parameters.]{
    \includegraphics[width=6.5cm]{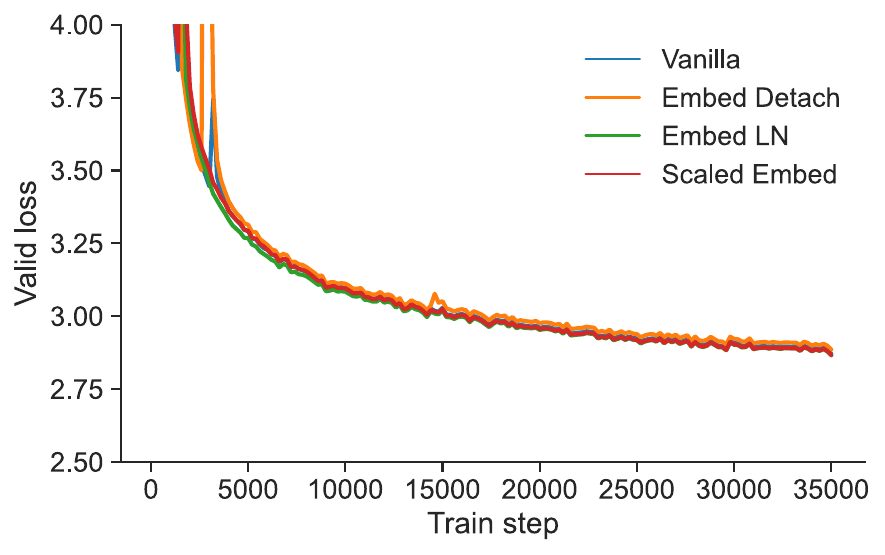}}
    \subfigure[1.7B parameters.]{
    \includegraphics[width=6.5cm]{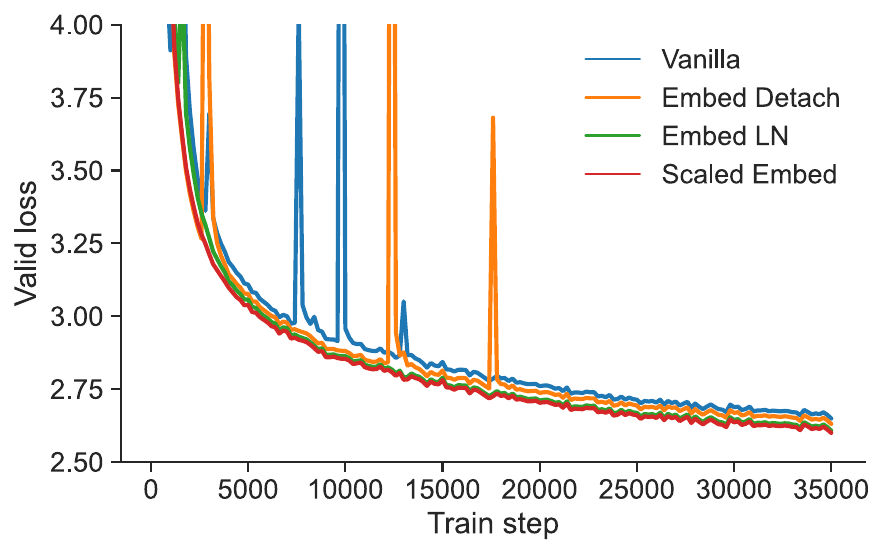}}
    \caption{Loss curves of each method in validation data.}
    \label{fig:valid_curve_comp}
\end{figure}

\begin{figure}[!t]
\centering
    \subfigure[350M parameters.]{
    \includegraphics[width=6.5cm]{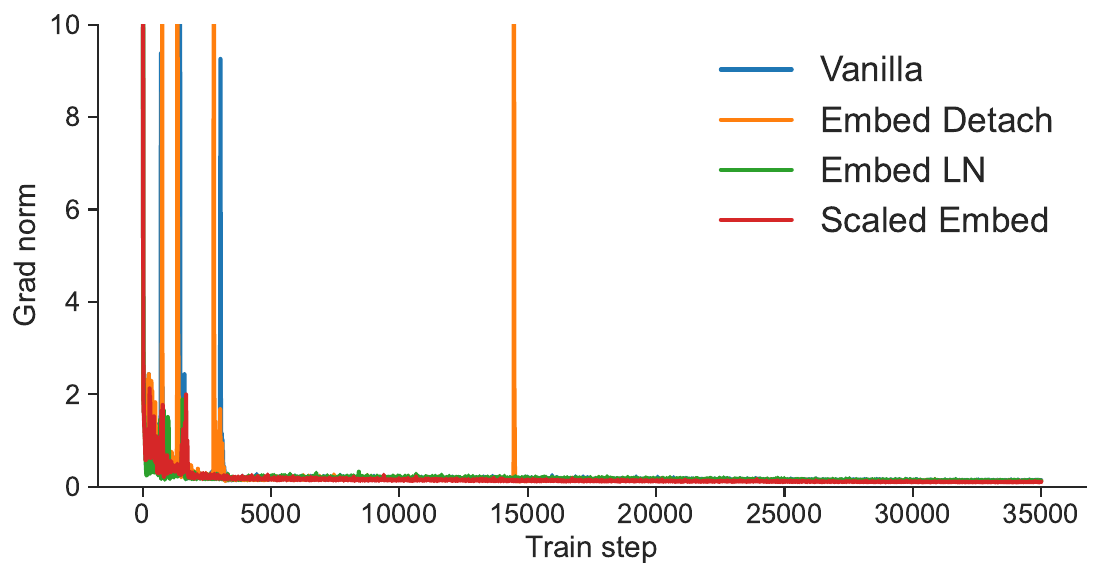}}
    \subfigure[1.7B parameters.]{
    \includegraphics[width=6.5cm]{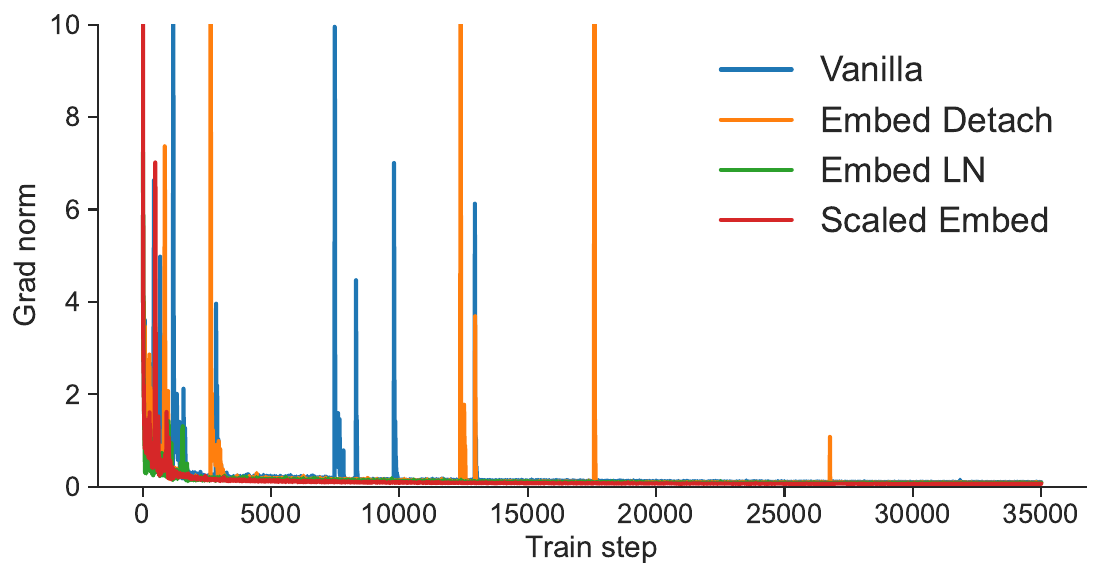}}
    \caption{Gradient norms of each method during the training.}
    \label{fig:grad_norm_comp}
\end{figure}

Figure \ref{fig:valid_curve_comp} shows the loss values of each method in validation data.
Figure \ref{fig:grad_norm_comp} shows the gradient norms of each method.
These figures indicate that Vanilla and Embed Detach faced loss and gradient spikes.
In contrast, Embed LN and Scaled Embed did not face spikes.
These results correspond to our theoretical analyses described in Sections \ref{sec:theory} and \ref{sec:proposed_method}.
Thus, only methods that make the upper bound of the gradient norm small have successfully avoided spikes in LLM pre-training.

\begin{wraptable}[14]{r}[0mm]{60mm}
  \vskip -13pt
  \centering
  \footnotesize
  \tabcolsep=1.5pt
  \begin{tabular}{ l | c c } \hline
  Model & WikiText $\downarrow$ & LAMBADA $\downarrow$ \\ \hline \hline
  \multicolumn{3}{c}{350M parameters} \\ \hline \hline
  Vanilla & 30.03 & 24.73 \\
  Embed Detach & 30.69 & 26.93 \\
  Embed LN & \textbf{29.85} & 25.03 \\
  Scaled Embed & 29.86 & \textbf{24.37} \\ \hline \hline
  \multicolumn{3}{c}{1.7B parameters} \\ \hline \hline
  Vanilla & 22.58 & 15.22 \\
  Embed Detach & 22.00 & 13.88 \\
  Embed LN & \textbf{21.29} & 13.00 \\
  Scaled Embed & \textbf{21.29} & \textbf{12.53} \\ \hline
  \end{tabular}
  \caption{Perplexities of each method.}
  \label{tab:ppl_in_main_exp}
\end{wraptable}

In comparison between 350M and 1.7B parameters, spikes occurred more frequently in 1.7B parameters.
Because we initialize embeddings with $\mathcal{N}(0, \sigma^2)$ where $\sigma = \sqrt{\frac{2}{5d}}$, the standard deviations of embeddings become small as $d$ gets larger in Vanilla and Embed Detach.
This means that the upper bounds described by Equations (\ref{eq:final_upper_bound_ffn}) and (\ref{eq:final_upper_bound_attn}) become large as $d$ gets larger because $\sigma_x$ and $\sigma_x'$ are nearly equal to the standard deviation of an input embedding in shallow layers.
Therefore, if we increase $d$ without any technique to control the upper bound of the gradient norms, a model becomes more unstable.
This result corresponds to the previous study reports~\citep{le-scao-etal-2022-language,chowdhery2022palm,zeng2023glmb} that their model became more unstable as they increased the number of parameters.

Table \ref{tab:ppl_in_main_exp} shows the perplexities of each method on WikiText and LAMBADA.
This table shows that Embed LN and Scaled Embed achieved comparable performance.
This result implies that methods have no significant difference from each other in their performance if each method prevents loss and gradient spikes.
In contrast, the perplexities of Vanilla and Embed Detach are worse except for Vanilla with 350M parameters in LAMBADA, and the difference in the performance is larger in a large amount of parameters.
This result implies that addressing spikes has a more serious influence on the performance as the parameter size gets larger.
We conduct experiments on a larger model in the following subsection.
Moreover, we discuss other settings in Appendix \ref{sec:discussions_on_other_configurations}.

\subsection{Results on 13B Parameter Model}
\label{sec:large_model}

We conduct experiments on pre-trainings of 13B parameter models to indicate that our provided conditions can stabilize a model with many more parameters than the ones discussed in Section \ref{sec:main_results}.
Due to the limitation of our computational budget, we focused on the comparison between Vanilla and Scaled Embed.
We tried two learning rates: $3.0 \times 10^{-4}$ and $1.0 \times 10^{-4}$.
Appendix \ref{sec:hyperparams} describes more details about hyper-parameters.

\begin{figure}[!t]
\centering
    \subfigure[$\mathrm{lr}=3.0 \times 10^{-4}$ .]{
    \includegraphics[width=6cm]{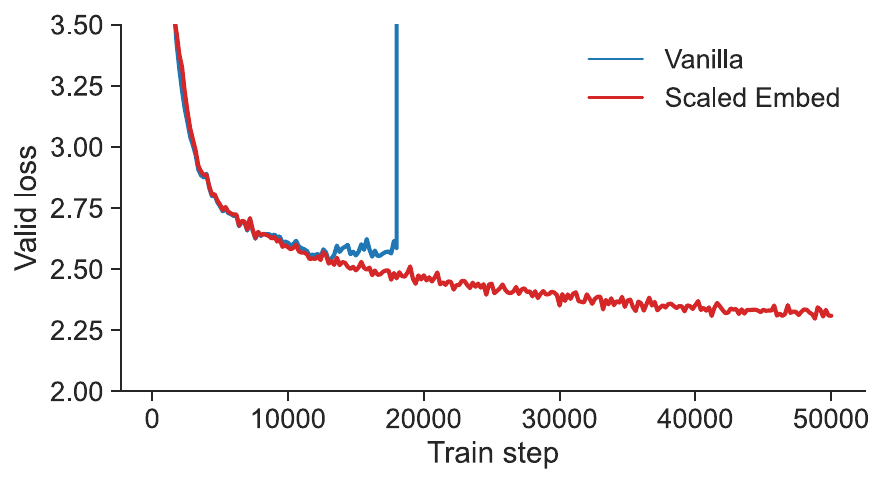}}
    \subfigure[$\mathrm{lr}=1.0 \times 10^{-4}$.]{
    \includegraphics[width=6cm]{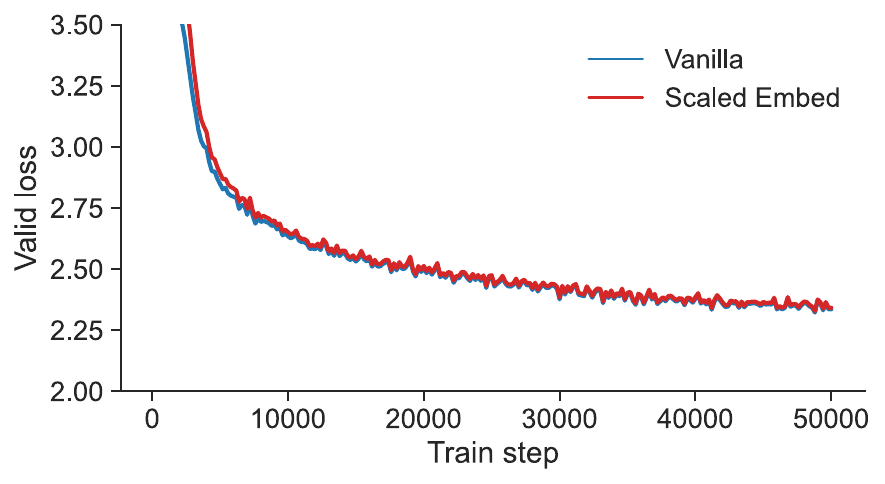}}
    \caption{Loss values of each method with 13B parameters when we use two learning rates: $\mathrm{lr}=3.0 \times 10^{-4}$ and $1.0 \times 10^{-4}$}
    \label{fig:13b_valid_curve}
\end{figure}

Figure \ref{fig:13b_valid_curve} shows the loss values of each configuration in validation data.
As shown in (a) of this figure, the loss value of Vanilla rose from approximately 10000 steps in $\mathrm{lr}=3.0 \times 10^{-4}$.
Then, the gradient of this model became too large to continue its pre-training.
In contrast, the loss value of Scaled Embed consistently decreased.
This result indicates that Scaled Embed stabilized the pre-training of the model with a large number of parameters.

\begin{table}[!t]
  \centering
  \footnotesize
  \begin{tabular}{ l | c c | c c} \hline
  & \multicolumn{2}{c|}{WikiText $\downarrow$} & \multicolumn{2}{c}{LAMBADA $\downarrow$} \\
  Model & $\mathrm{lr}=3.0 \times 10^{-4}$ & $\mathrm{lr}=1.0 \times 10^{-4} $ & $\mathrm{lr}=3.0 \times 10^{-4}$& $\mathrm{lr}=1.0 \times 10^{-4}$ \\ \hline
  Vanilla & N/A & 15.12 & N/A & 6.50 \\
  Scaled Embed & \textbf{14.47} & 15.25 & \textbf{5.97} & 6.53 \\ \hline
  \end{tabular}
  \caption{Perplexities of each method with 13B parameters when we use two learning rates: $\mathrm{lr}=3.0 \times 10^{-4}$ and $1.0 \times 10^{-4}$.}
  \label{tab:13b_ppl}
\end{table}

\begin{table}[!t]
  \centering
  \footnotesize
  \tabcolsep=1.5pt
  \begin{tabular}{ l | c c c c c c} \hline
  Model & \multicolumn{1}{c|}{PIQA $\uparrow$} & \multicolumn{1}{c|}{OpenBookQA $\uparrow$}
  & \multicolumn{1}{c|}{HellaSwag $\uparrow$} & \multicolumn{1}{c|}{WinoGrande $\uparrow$}
  & \multicolumn{1}{c|}{ARC-easy $\uparrow$} & \multicolumn{1}{c}{ARC-challenge $\uparrow$}\\ \hline \hline
  \multicolumn{1}{c}{}& \multicolumn{6}{c}{$\mathrm{lr}=3.0 \times 10^{-4}$}\\ \hline \hline
  Vanilla & \multicolumn{6}{c}{N/A} \\
  Scaled Embed & \textbf{78.94} & \textbf{39.20} & \textbf{71.03} & \textbf{63.77} & \textbf{60.31} & \textbf{35.49} \\ \hline \hline
  \multicolumn{1}{c}{}& \multicolumn{6}{c}{$\mathrm{lr}=1.0 \times 10^{-4}$}\\ \hline \hline
  Vanilla & 77.80 & 38.40 & 69.10 & 61.64 & 57.95 & 33.53 \\
  Scaled Embed & 78.07 & \textbf{39.20} & 68.92 & 62.59 & 58.92 & 33.19 \\ \hline
  \end{tabular}
  \caption{Performance of each method with 13B parameters on standard tasks.}
  \label{tab:13b_benchmark}
\end{table}

Table \ref{tab:13b_ppl} shows the perplexities of each configuration in evaluation data.
This table indicates that we can achieve better performance when we use a larger learning rate.
In addition, the perplexities of Scaled Embed were comparable to ones of Vanilla when we used the small learning rate: $\mathrm{lr}=1.0 \times 10^{-4}$.
Table \ref{tab:13b_benchmark} shows the performance of each configuration on the standard benchmark datasets: PIQA~\citep{piqa2020}, OpenBookQA~\citep{mihaylov-etal-2018-suit}, HellaSwag~\citep{zellers-etal-2019-hellaswag}, WinoGrande~\citep{10.1145/3474381} and ARC easy and challenge~\citep{clark2018thinksolvedquestionanswering}.
This table shows results consistent with the evaluation in terms of perplexity.
These results imply that our provided conditions have no considerable risk in pre-training.
Thus, we have to satisfy large shortcut in addition to small sub-layers to stabilize the pre-trainings of LLMs.

\section{Related Work}
\label{sec:related_work}
\paragraph{Stability}
To stabilize trainings of Transformer-based neural language models, there have been various discussions on the architecture~\citep{DBLP:conf/icml/XiongYHZZXZLWL20,liu-etal-2020-understanding,takase-etal-2023-b2t,zeng2023glmb,stabilizing-transformer-training}, initialization method~\citep{nguyen-salazar-2019-transformers,conf/iclr/ZhangDM19,pmlr-v119-huang20f,https://doi.org/10.48550/arxiv.2203.00555}, training strategy~\citep{zhang2022opt,NEURIPS2022_aac02401}, and loss function~\citep{chowdhery2022palm,wortsman2023smallscale}.

\citet{DBLP:conf/icml/XiongYHZZXZLWL20} theoretically analyzed gradient scales of each part in Transformers, and indicated that the Pre-LN Transformer is more stable than the Post-LN Transformer, that is the original Transformer architecture~\citep{NIPS2017_7181}.
Since the Pre-LN Transformer is more stable than the Post-LN Transformer theoretically and empirically, recent studies mainly have used the Pre-LN Transformer to construct an LLM.
We also assume using the Pre-LN Transformer in the analysis on the training dynamics in this paper.

To stabilize the LLM pre-training, \citet{le-scao-etal-2022-language} applied the layer normalization to the embedding layer.
\citet{zeng2023glmb} used shrink embedding gradient technique~\citep{NEURIPS2021_a4d92e2c}.
\citet{nishida-etal-2024-initialization} proposed weight scaling as reparameterization (WeSaR) which uses additional parameters to scale parameters of internal layers.
In this study, we theoretically proved that the layer normalization to the embedding layer controls the upper bound of the gradient norms of sub-layers when we use the widely used initialization method for LLMs~\citep{nguyen-salazar-2019-transformers,shoeybi2020megatronlm}, and thus, it stabilizes the pre-training.

For the initialization methods, recent studies have proposed maximal update parameterization ($\mu$P) and its variants as approaches to transfer hyper-parameters from small models to larger ones without incurring the cost of hyper-parameter search~\cite{mupinicml,yang2024spectralconditionfeaturelearning,yang2024tensor}.
However, as described in \citet{yang2024tensor}, such methods may be ineffective when applied to Transformers, as their underlying assumptions do not accurately reflect the actual Transformer architectures.
In studies focusing on Transformers, \citet{nguyen-salazar-2019-transformers} proposed a strategy to initialize parameters of Transformers with small values to stabilize their training.
\citet{conf/iclr/ZhangDM19} and \citet{pmlr-v119-huang20f} indicated that we can remove layer normalizations in Transformers if we use their proposed initialization methods.
\citet{https://doi.org/10.48550/arxiv.2203.00555} adjusted initial parameter scales based on the number of layers to stabilize the Post-LN Transformer.
In this study, we indicated that the widely used initialization method~\citep{shoeybi2020megatronlm}, which makes parameters small, is necessary to stabilize the LLM pre-training.
Moreover, we proved that we can prevent the loss spike problem by making the standard deviation of embeddings close to $1$.

\paragraph{Efficiency}
As shown in Section \ref{sec:large_model} and Appendix \ref{sec:other_lr}, our modification enables the pre-training with a relatively larger learning rate, and can achieve better performance.
Thus, this study can be regarded as on the efficiency of LLM pre-training because our modification can construct a better LLM with a given budget.
\citet{strubell-etal-2019-energy} and \citet{DBLP:journals/corr/abs-1907-10597} reported that recent neural methods require substantial computational costs, and thus, they argued that we have to explore a cost-efficient approach.
%We can split paragraph here if we need
\citet{rajbhandari2020zero} proposed ZeRO that reduces memory redundancies during the multi GPU training without increasing communication volume.
\citet{dao2022flashattention} focused on GPU memory reads/writes, and proposed FlashAttention that accelerates the speed of attention mechanisms in Transformers.
To reduce the number of computations in the attention mechanism, \citet{shazeer2019fast} proposed the multi-query attention that shares one key and value across all of the attention heads in each layer.
\citet{takase-kiyono-2023-lessons} explored several parameter sharing strategies, and indicated that parameter sharing across some layers can achieve comparable performance to the vanilla model with a small number of parameters.
Moreover, several studies have explored a better construction way with a limited budget~\citep{izsak-etal-2021-train,takase-kiyono-2021-rethinking}.
We believe that we can take advantage of their findings to make our LLMs more efficient.

\section{Conclusion}
This paper explored why large language models (LLMs) sometimes experience loss spikes during pre-training.
To provide evidence, we specifically focused on the gradients of sub-layers.
We introduced an upper bound for the gradient norms through an analysis of the spectral norms of the Jacobian matrices for the sub-layers.
We then theoretically identified two conditions for avoiding loss spikes: small sub-layers and large shortcut.
To meet these conditions, we show that using the widely adopted initialization method for LLMs can make the sub-layer parameters small, and that embedding scaling or incorporating layer normalization into the embedding layer can make the standard deviation of each embedding close to $1$, resulting in large shortcut.
Experimental results indicated that methods satisfying these conditions avoid loss spikes.
Furthermore, these methods allow for training with a relatively larger learning rate, leading to improved performance.
We hope our theoretical analyses and empirical findings will help avoid wasting valuable time and computational budgets during LLM construction.

\section*{Ethics Statement}
To stabilize the LLM pre-training, this paper provides theoretical analyses on the spectral norms of the Jacobian matrices for sub-layers to estimate the upper bound of the gradient norm of $\mathcal{L}$.
This paper focuses on only the stability of LLM pre-training, and thus, we have to address other issues of LLMs such as hallucinations to use the LLM in a real application.

\section*{Reproducibility Statement}
We do not aim to propose a novel method in this paper, but we mainly focus on theoretical analyses on the spectral norms of the Jacobian matrices to find the factor to stabilize the pre-training of LLMs.
We justify our theoretical analyses through experiments with various situations.
To activate our modification, we add only several lines to a widely used implementation, i.e., Megatron-LM\footnote{\href{https://github.com/NVIDIA/Megatron-LM}{https://github.com/NVIDIA/Megatron-LM}}.
Therefore, we believe that it is easy to reproduce our experimental results.
Moreover, we did not modify the internal architecture of Transformers.
In particular, Scaled Embed only scales embeddings by a constant value, making it convenient in terms of portability across various implementations, such as Transformers in HuggingFace\footnote{\href{https://huggingface.co/docs/transformers/main/index}{https://huggingface.co/docs/transformers/main/index}} and vLLM\footnote{\href{https://github.com/vllm-project/vllm}{https://github.com/vllm-project/vllm}}.

However, because it is difficult to conclude that our provided conditions completely solve the instability during pre-training of LLMs, it is better to combine other techniques to stabilize the pre-training such as an auxiliary loss described by \citet{chowdhery2022palm} to make the pre-training more stable in an actual pre-training situation.

% \bibliography{reference}
\bibliographystyle{colm2025_conference}

% \newpage
\appendix
\section{Hyper-Parameters}
\label{sec:hyperparams}

Table \ref{tab:hyper_params} shows that hyper-parameters used in our experiments on LLMs.
To make the experiments on 13B parameter models close to a realistic situation, we increased the batch size and the number of updates in comparison to ones in experiments on 350M and 1.7B parameter models.
In addiiton, for Adam $\beta_2$, most studies have used $0.95$ to stabilize their pre-trainings~\citep{NEURIPS2020_1457c0d6,zhang2022opt,zeng2023glmb,10.5555/3618408.3618510,touvron2023llama}, and thus, we also used $0.95$ in the experiments on 13B parameter models.
As described in Section \ref{sec:large_model}, we emphasize that the pre-training of Vanilla is essentially unstable even if we use the widely used Adam $\beta_2$ value, $0.95$, which is known as the technique to stabilize the pre-training.
Therefore, our provided conditions are also effective for the stabilization in the realistic situation.

\section{Discussions on Other Configurations}
\label{sec:discussions_on_other_configurations}
In this section, we conduct experiments on other configurations to describe connections with previous study reports.

\subsection{Methods without Small Sub-Layers}
\label{sec:xavier_init}
Since we applied the widely used initialization method for LLMs in experiments in Section \ref{sec:main_exp}, all methods satisfy the condition on the small sub-layers.
In this section, we empirically investigate the property of the method that violates the condition.
We compare the Transformer initialized by the Xavier initialization~\citep{pmlr-v9-glorot10a} (Xavier Init), and the combination of Xavier Init and Scaled Embed (Xavier Init + Scaled Embed) with Scaled Embed.
As described in Table \ref{tab:method_property}, Xavier Init violates both conditions, and Xavier Init + Scaled Embed satisfies the only large shortcut condition.
In the same manner as in Section \ref{sec:main_exp}, we trained models of 350M and 1.7B parameters with $\mathrm{lr} = 5.0 \times 10^{-4}$.
We also used the hyper-parameters described in Table \ref{tab:hyper_params}.

\begin{table}[!t]
  \centering
  \caption{Hyper-parameters used in our experiments on the LLM pre-training.}
  \label{tab:hyper_params}
  \begin{tabular}{ l | c c c} \hline
  Name & 350M & 1.7B & 13B\\ \hline
  Layer num & 24 & 24 & 40\\
  Hidden dim size & 1024 & 2304 & 5120\\
  FFN dim size & 4096 & 9216 & 20480\\
  Attention heads & 16 & 24 & 40\\
  Dropout rate & 0.1 & 0.1 & 0.1\\
  Precision & \texttt{float16} & \texttt{float16} & \texttt{float16} \\
  Sequence length & 2048 & 2048 & 2048\\
  Batch size & 528 & 528 & 1024 \\
  The number of updates & 35000 & 35000 & 50000 \\
  Adam $\beta_1$ & 0.9 & 0.9 & 0.9\\
  Adam $\beta_2$ & 0.999 & 0.999 & 0.95 \\
  Gradient clipping & 1.0 & 1.0 & 1.0\\
  $lr$ decay style & cosine & cosine & cosine \\
  $lr$ warmup fraction & 0.05 & 0.05 & 0.05\\
  Weight decay & 0.01 & 0.01 & 0.01\\ \hline
  \end{tabular}
\end{table}

\begin{table}[!t]
  \centering
  \footnotesize
  \tabcolsep=1.5pt
  \begin{tabular}{ l | c c } \hline
  Method & Small sub-layers & Large shortcut \\ \hline \hline
  Xavier Init & - & - \\
  Vanilla & $\checkmark$ & - \\
  Embed Detach & $\checkmark$ & - \\
  Embed LN & $\checkmark$ & $\checkmark$ \\
  Scaled Embed & $\checkmark$ & $\checkmark$ \\
  Xavier Init + Scaled Embed & - & $\checkmark$ \\ \hline
  \end{tabular}
  \caption{Relations between each method in experiments and two conditions to control the upper bound of gradient norms.}
  \label{tab:method_property}
\end{table}

\begin{table}[!t]
  \centering
  \footnotesize
  \tabcolsep=1.5pt
  \begin{tabular}{ l | c c } \hline
  Model & WikiText $\downarrow$ & LAMBADA $\downarrow$ \\ \hline \hline
  \multicolumn{3}{c}{350M parameters} \\ \hline \hline
  Xavier Init & 33.92 & 34.72 \\
  Xavier Init + Scaled Embed & 30.50 & 26.55 \\
  Scaled Embed & \textbf{29.86} & \textbf{24.37} \\ \hline \hline
  \multicolumn{3}{c}{1.7B parameters} \\ \hline \hline
  Xavier Init & 30.10 & 29.29 \\
  Xavier Init + Scaled Embed & 23.16 & 15.49 \\
  Scaled Embed & \textbf{21.29} & \textbf{12.53} \\ \hline
  \end{tabular}
  \caption{Perplexities of each method.}
  \label{tab:ppl_xavier_init}
\end{table}

\begin{figure}[!t]
\centering
    \subfigure[350M parameters.]{
    \includegraphics[width=6cm]{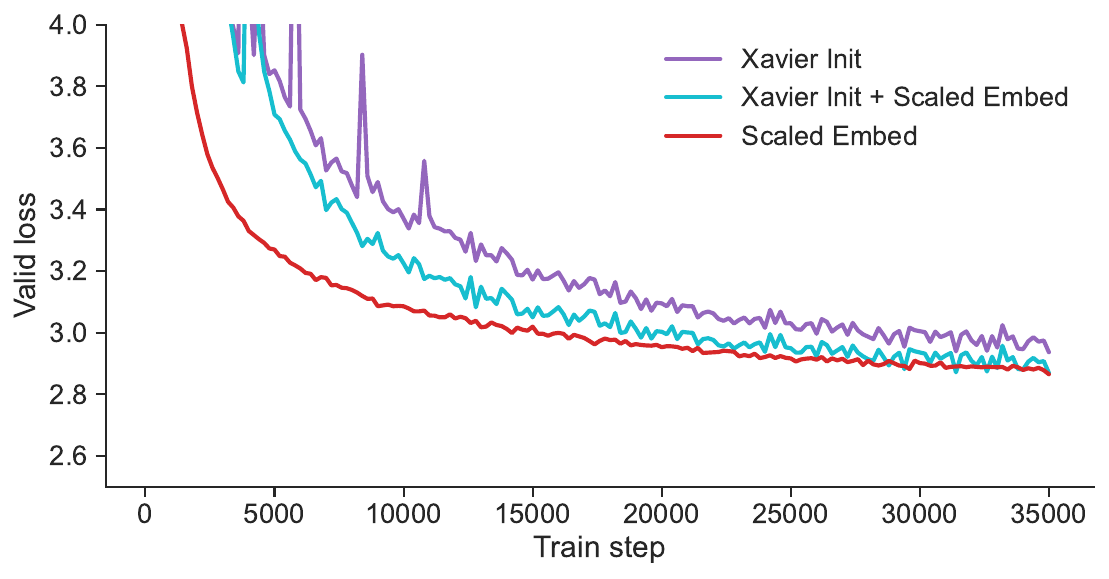}}
    \subfigure[1.7B parameters.]{
    \includegraphics[width=6cm]{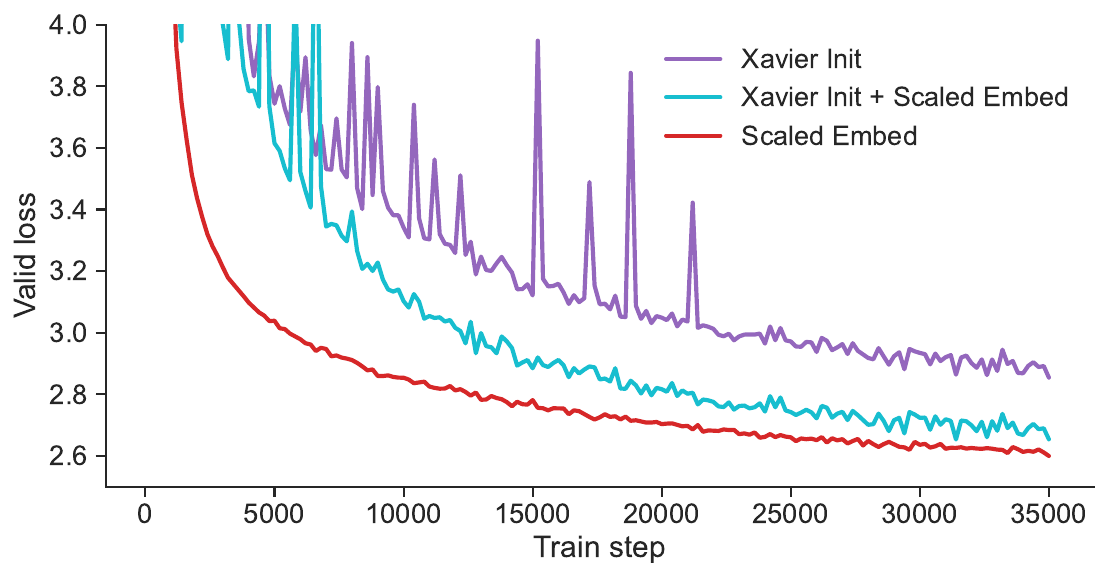}}
    \caption{Loss curves of each method in validation data for the comparison to methods without small sub-layers.}
    \label{fig:valid_curve_xavier_init}
\end{figure}

\begin{figure}[!t]
\centering
    \subfigure[350M parameters.]{
    \includegraphics[width=6cm]{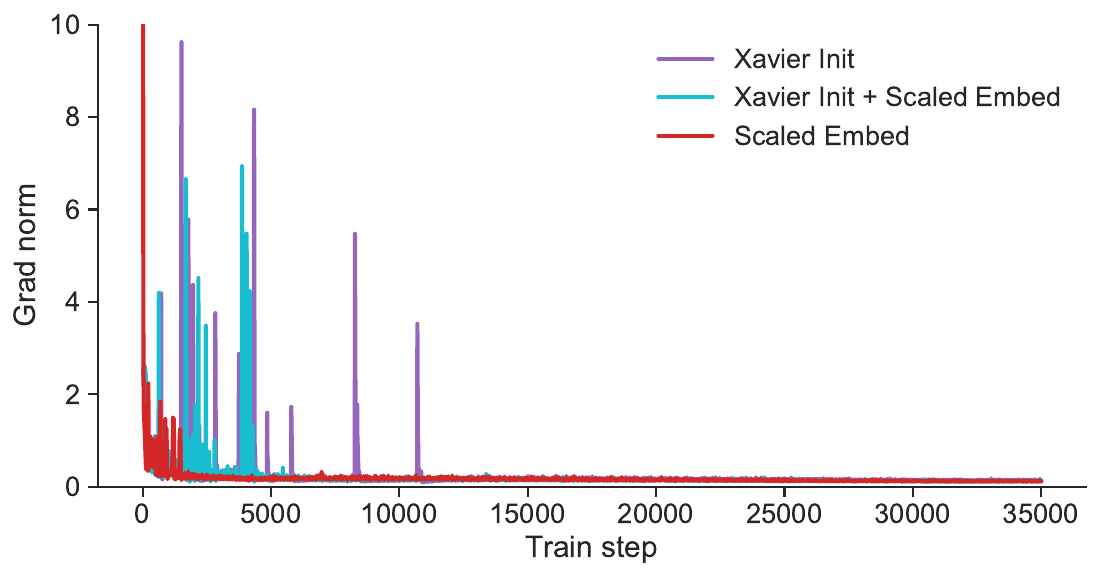}}
    \subfigure[1.7B parameters.]{
    \includegraphics[width=6cm]{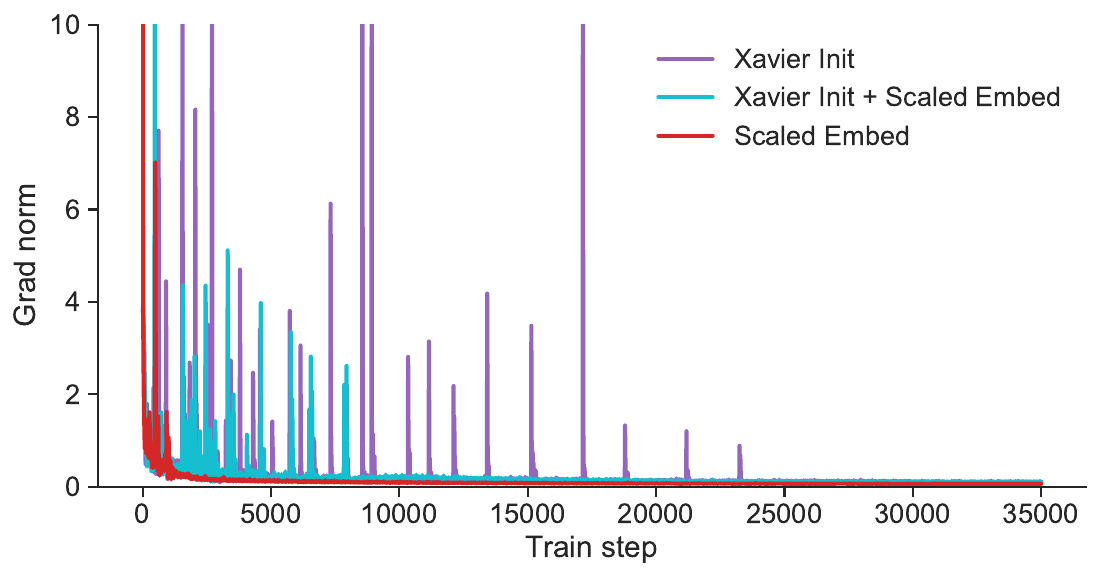}}
    \caption{Gradient norms of each method during the training for the comparison to methods without small sub-layers.}
    \label{fig:gradnorm_xavier_init}
\end{figure}

Figure \ref{fig:valid_curve_xavier_init} shows loss curves in validation data for 350M and 1.7B parameters in each method, and Figure \ref{fig:gradnorm_xavier_init} shows their gradient norms.
These figures show that Xavier Init and Xavier Init + Scaled Embed faced loss and gradient spikes.
In particular, the spikes appeared more frequently in Xavier Init, which violates both conditions, in comparison with Xavier Init + Scaled Embed.
In contrast, Scaled Embed, which satisfies both conditions, avoided the gradient spike and prevented the loss spike problem.
These results indicate that we have to satisfy both conditions: small sub-layers and large shortcut to prevent the loss spike problem.
Moreover, Table \ref{tab:ppl_xavier_init} shows perplexities of each configuration in evaluation data.
This table indicates that Scaled Embed achieved better performance than Xavier Init and Xavier Init + Scaled Embed that faced some spikes.

\subsection{Varying Learning Rate}
\label{sec:other_lr}

\citet{le-scao-etal-2022-language} reported that the stable method, such as Embed LN, was worse than Vanilla.
However, in Section \ref{sec:main_results}, the stable methods, Scaled Embed and Embed LN, achieved better performance than Vanilla in the 1.7B parameter configuration.
We suppose that the difference in the learning rate causes this gap in findings.
In this section, we tried to train Vanilla and Scaled Embed with larger and smaller learning rates: $\mathrm{lr}=1.0 \times 10^{-3}$ and $1.0 \times 10^{-4}$ respectively.

\begin{figure}[!t]
\centering
    \subfigure[350M with $\mathrm{lr}=1.0 \times 10^{-3}$.]{
    \includegraphics[width=4.5cm]{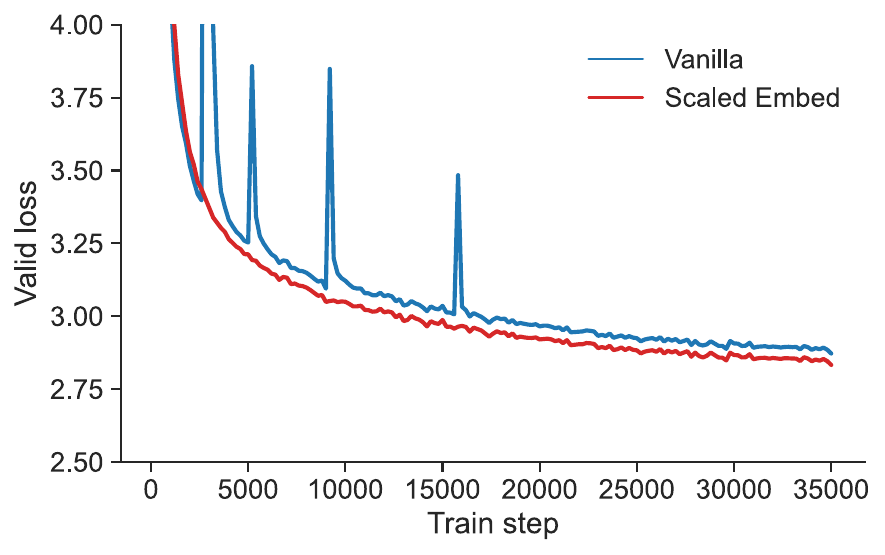}}
    % \caption{350M parameters.}
    \subfigure[350M with $\mathrm{lr}=5.0 \times 10^{-4}$.]{
    \includegraphics[width=4.5cm]{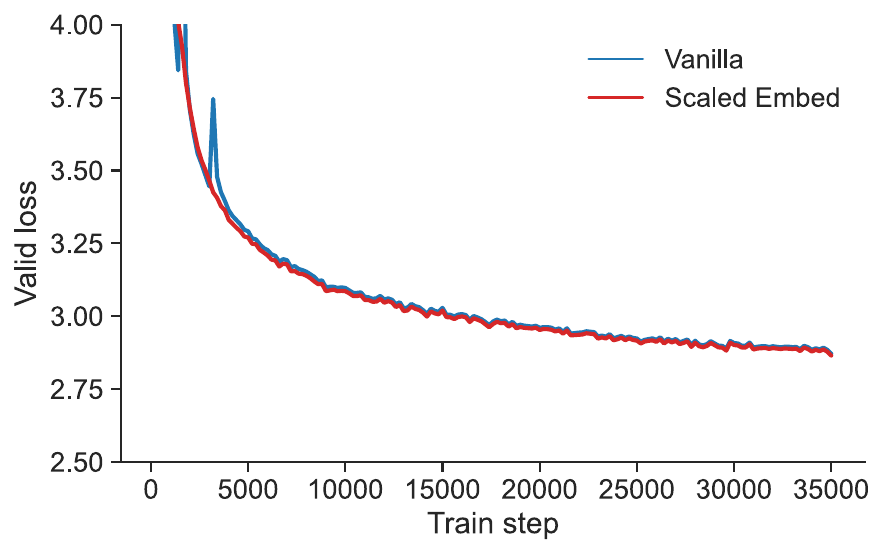}}
    \subfigure[350M with $\mathrm{lr}=1.0 \times 10^{-4}$.]{
    \includegraphics[width=4.5cm]{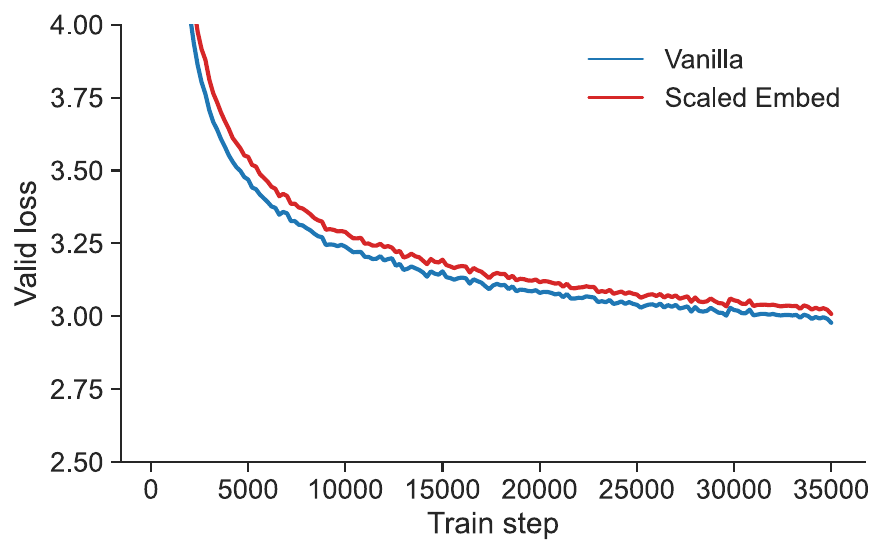}}
    \subfigure[1.7B with $\mathrm{lr}=1.0 \times 10^{-3}$.]{
    \includegraphics[width=4.5cm]{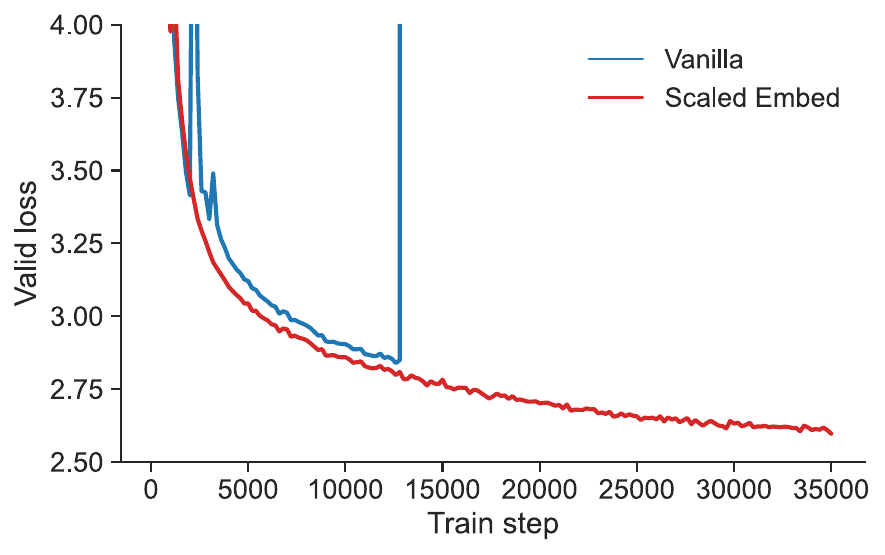}}
    \subfigure[1.7B with $\mathrm{lr}=5.0 \times 10^{-4}$.]{
    \includegraphics[width=4.5cm]{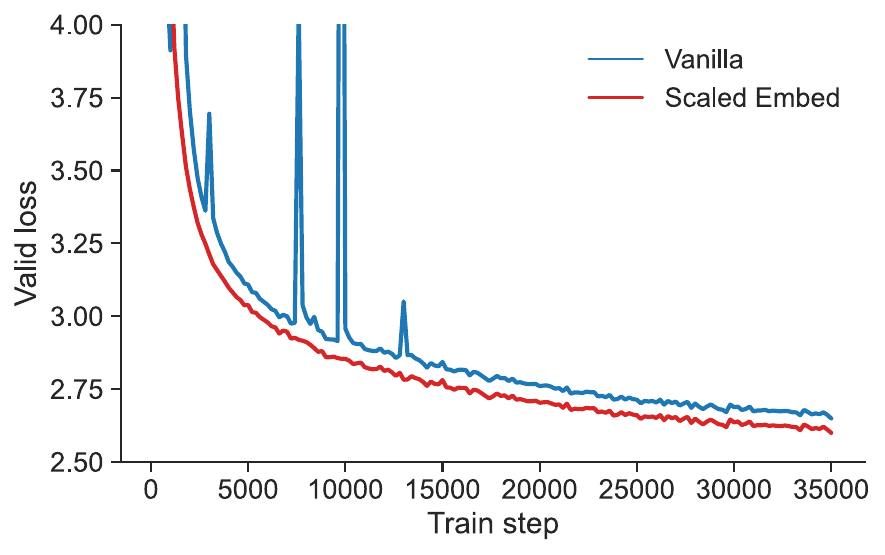}}
    \subfigure[1.7B with $\mathrm{lr}=1.0 \times 10^{-4}$.]{
    \includegraphics[width=4.5cm]{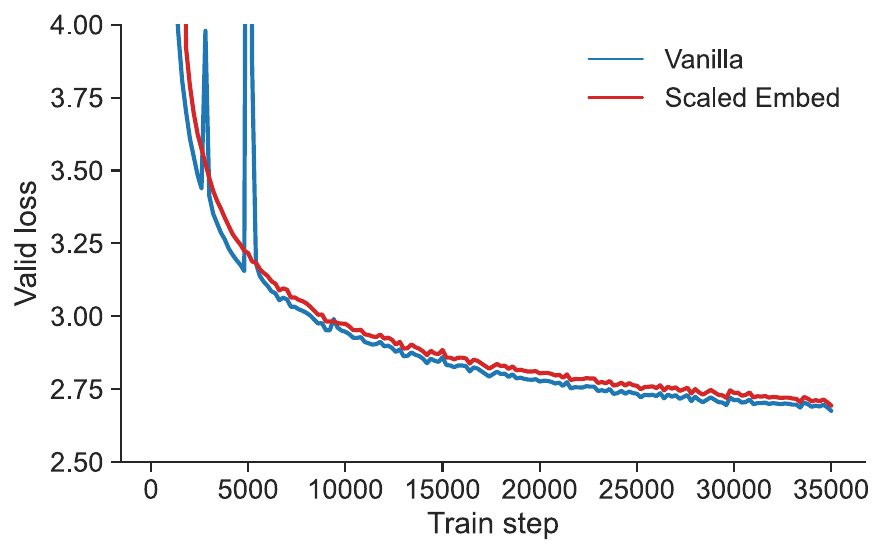}}
    \caption{Loss values of each method with 350M and 1.7B parameters when we vary a learning rate.}
    \label{fig:vary_lr}
\end{figure}

\begin{table}[!t]
  \centering
  \tabcolsep=1.5pt
  \footnotesize
  % \scriptsize
  \begin{tabular}{ l | c c c | c c c} \hline
  & \multicolumn{3}{c|}{WikiText $\downarrow$} & \multicolumn{3}{c}{LAMBADA $\downarrow$} \\
  Model & $\mathrm{lr}\ 1.0 \times 10^{-3}$ & $\mathrm{lr}\ 5.0 \times 10^{-4}$ & $\mathrm{lr}\ 1.0 \times 10^{-4}$ & $\mathrm{lr}\ 1.0 \times 10^{-3}$ & $\mathrm{lr}\ 5.0 \times 10^{-4}$ & $\mathrm{lr}\ 1.0 \times 10^{-4}$ \\ \hline \hline
  \multicolumn{7}{c}{350M parameters} \\ \hline \hline
  Vanilla & 29.96 & 30.35 & 34.51 & 25.12 & 24.73 & 32.49 \\
  Scaled Embed & \textbf{28.09} & 29.86 & 35.66 & \textbf{22.03} & 24.37 & 37.14 \\ \hline \hline
  \multicolumn{7}{c}{1.7B parameters} \\ \hline \hline
  Vanilla & N/A & 22.58 & 23.54 & N/A & 15.22 & 16.17\\
  Scaled Embed & \textbf{20.95} & 21.29 & 23.78 & \textbf{12.26} & 12.53 & 15.39 \\ \hline
  \end{tabular}
  \caption{Perplexities of each method with 350M and 1.7B parameters when we vary a learning rate.}
  \label{tab:ppl_in_vary_lr}
\end{table}

Figure \ref{fig:vary_lr} shows loss values of each configuration in validation data.
As shown in this figure, the larger the learning rate we used, the more frequent the spikes occurred in Vanilla.
In particular, in $\mathrm{lr}=1.0 \times 10^{-3}$, the training of Vanilla with 1.7B parameters failed because its gradient exploded.
In contrast, Scaled Embed stabilized the training, and thus, its loss values consistently decreased.

Table \ref{tab:ppl_in_vary_lr} shows the perplexities of each configuration in evaluation data.
This table indicates that Vanilla with 350M parameters achieved better performance in $\mathrm{lr}=1.0 \times 10^{-4}$ that is the situation where its training did not face any spike.
This result corresponds to the report of \citet{le-scao-etal-2022-language}.
Thus, we suppose that they conducted the comparison with a too-small learning rate to stabilize Vanilla.
In contrast, the stable methods are more effective in training with a large learning rate, as shown in Figure \ref{fig:vary_lr} and Table \ref{tab:ppl_in_vary_lr}.
Therefore, if \citet{le-scao-etal-2022-language} used a relatively large learning rate in their experiments, their stable method could achieve better performance.

\subsection{Varying Sequence Length}
\label{sec:other_seqlen}

\citet{NEURIPS2022_aac02401} indicated that it is better to train with a short sequence at the early stage to stabilize the LLM pre-training.
They justified their method based on the curriculum learning strategy.
On the other hand, in this section, we provide the theoretical justification to their method in terms of the standard deviation of the shortcut.

\begin{wrapfigure}[18]{r}[0pt]{6cm}
  \vskip -0.1pt
  \centering 
    \includegraphics[width=6cm]{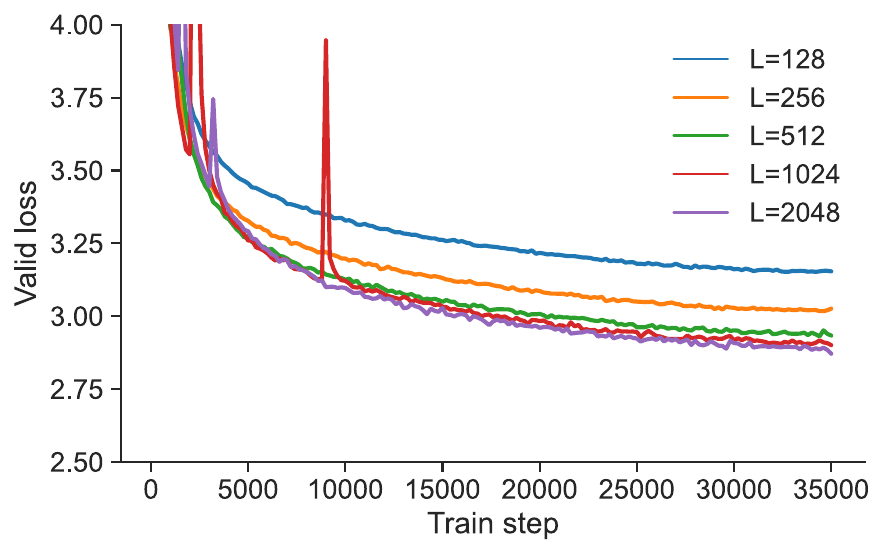}
   \caption{Loss curves of Vanilla with 350M parameters in validation data when we vary the input sequence length. We adjust the batch size to use the same number of tokens for the training of each model.}
   \label{fig:vary_seqlen}
\end{wrapfigure}

As described in Section \ref{sec:preln}, Transformers add the output of each sub-layer to the shortcut.
Since the standard deviation of the self-attention layer tends to decrease with the length of an input sequence especially at the early stage\footnote{See Appendix \ref{sec:std_selfattn} for details.}, a long sequence tends to keep the standard deviation of the shortcut small.
Therefore, the long sequence makes the pre-training of Vanilla more unstable.

We conducted experiments with varying the length of the input sequence $L$ from $128$ to $2048$.
To use the same number of tokens to update parameters, we adjusted the batch size.
Figure \ref{fig:vary_seqlen} shows loss values of Vanilla with each $L$ configuration in the validation data.
This figure shows that spikes occurred only in the large $L$, i.e., $1024$ and $2048$.
Moreover, the spikes are more likely to occur at the early stage of the pre-training.
Therefore, using a short sequence stabilizes the training at the early stage, as reported in \citet{NEURIPS2022_aac02401}.

\subsection{RMSNorm}
\label{sec:rmsnorm}
Some recent LLMs use the RMSNorm~\citep{NEURIPS2019_1e8a1942} instead of the LN in their Transformers~\citep{touvron2023llama}.
We discuss such an architecture in this section.
In the same as LN discussed in Section \ref{sec:analysis_on_jacobian_ffn}, we can obtain the Jacobian matrix of the RMSNorm with the following equation:
\begin{align}
\bigg\| \frac{\partial \mathrm{RMSNorm}(x)}{\partial x}\bigg\|_2 = \frac{1}{\sigma_x}I \label{eq:_rmsnorm_grad}
\end{align}
Thus, the upper bound of the gradient norm is the same in LN if we use RMSNorm.

\begin{figure*}[!t]
\centering
    \subfigure[Loss curves in validation data.]{
    \includegraphics[width=6cm]{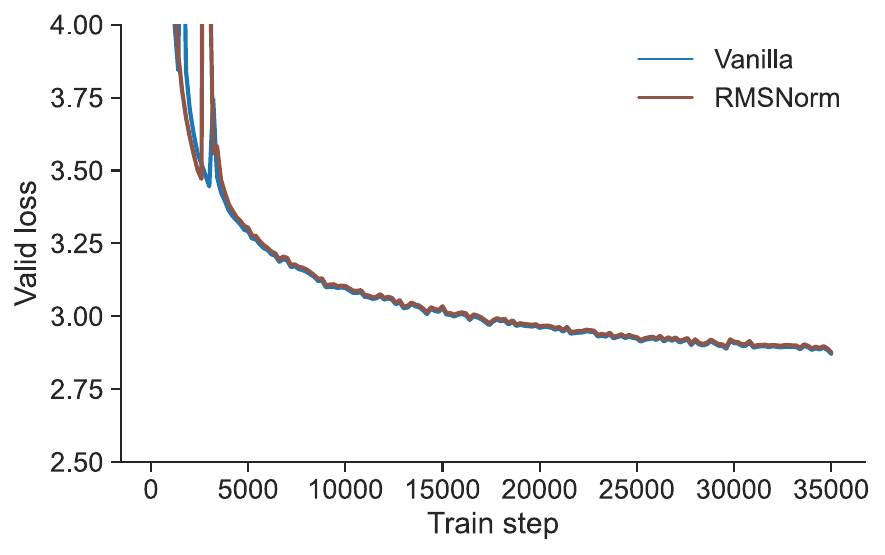}}
    \subfigure[Gradient norms.]{
    \includegraphics[width=6cm]{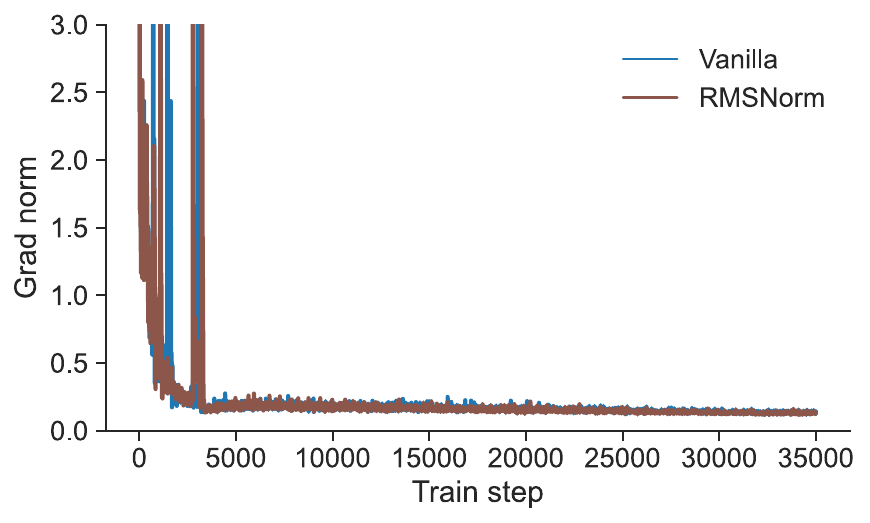}}
    \caption{Loss values and gradient norms of Vanilla and RMSNorm.}
    \label{fig:result_rmsnorm}
\end{figure*}

Figure \ref{fig:result_rmsnorm} shows the loss values and gradient norms of the Vanilla configuration in Section \ref{sec:main_results} and the one using RMSNorms instead of LNs (``RMSNorm'' in figures) with 350M parameters.
We trained them with $\mathrm{lr}= 5.0 \times 10^{-4}$ as in Section \ref{sec:main_results}\footnote{We tried to train them with $\mathrm{lr}= 1.0 \times 10^{-3}$ but RMSNorm exploded.}.
As shown in these figures, RMSNorm faced loss and gradient spikes in a similar location to the ones of Vanilla.
These empirical results also indicate that the RMSNorms have the same problem as LNs regarding the instability.

\section{Scaling Embeddings with Larger Value}
\label{sec:scaling_larger_value}

Equations (\ref{eq:final_upper_bound_ffn}) and (\ref{eq:final_upper_bound_attn}) indicate that we can stabilize the LLM pre-training by adjusting the standard deviation of the shortcut to a large value.
In fact, our experimental results show that we can stabilize the LLM pre-training by making the standard deviation of each embedding close to $1$.
To investigate how about a larger value, we conducted experiments with making the standard deviation of each embedding close to 5 and 50.

\begin{figure*}[!t]
\centering
    \subfigure[350M parameters.]{
    \includegraphics[width=6cm]{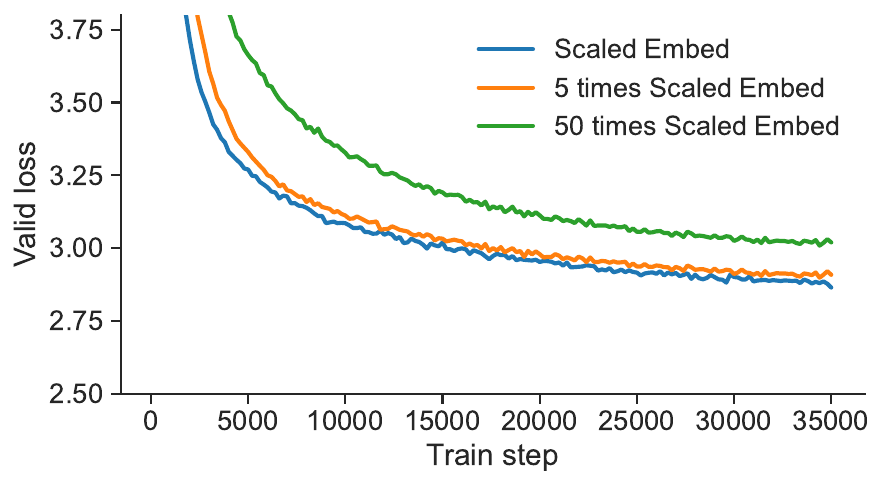}}
    \subfigure[1.7B parameters.]{
    \includegraphics[width=6cm]{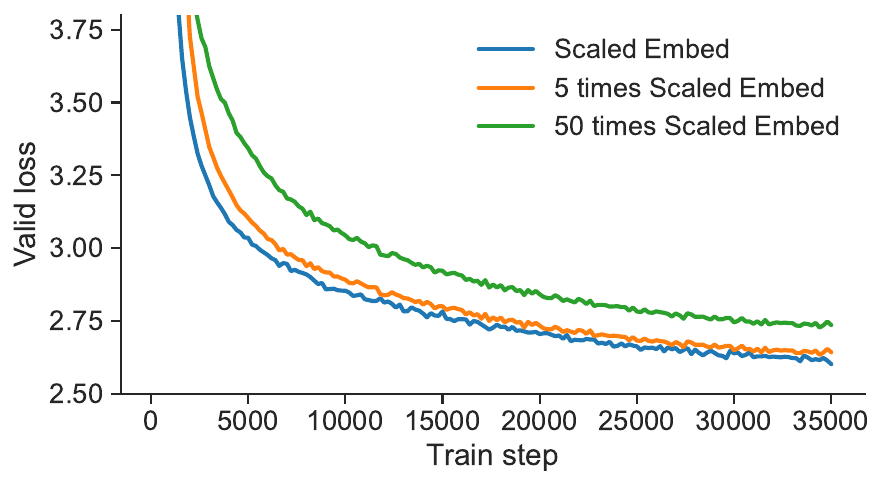}}
    \caption{Loss curves in validation data when we scale the standard deviation of embeddings with larger than $1$.}
    \label{fig:scaling_larger_value}
\end{figure*}

Figure \ref{fig:scaling_larger_value} shows loss curves in validation data for 350M and 1.7B parameters in each situation.
This figure indicates that although all settings prevented the loss spike problem, the larger standard deviation than $1$ degraded the performance.
Therefore, it is unnecessary to scale the standard deviation of each embedding with a larger value than $1$ to prevent the performance degradation.

\section{Distributions of Sub-Layer Outputs}
\label{sec:sublayer_distribution}

\begin{figure}[!t]
\centering
    \subfigure[self-attention of 350M parameters.]{
    \includegraphics[width=4.5cm]{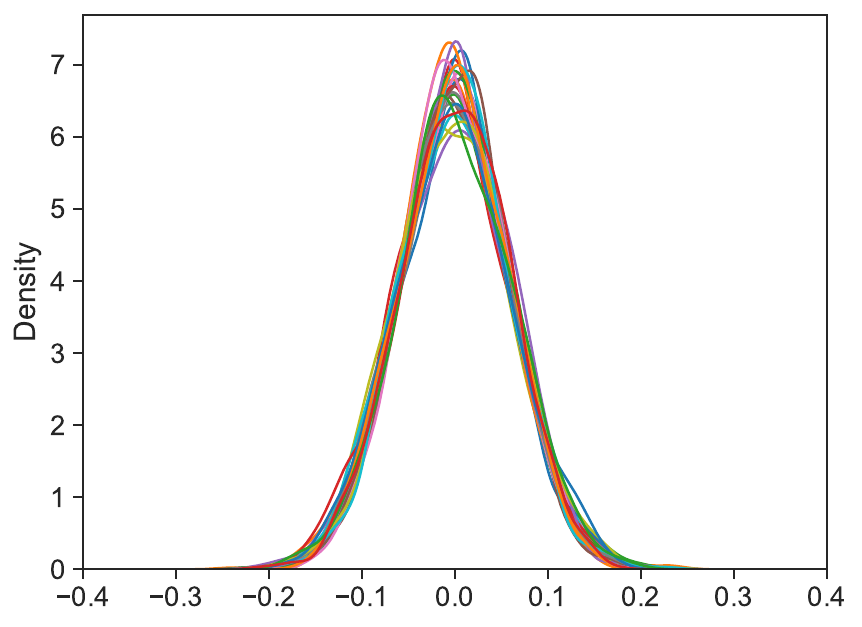}}
    \subfigure[self-attention of 1.7B parameters.]{
    \includegraphics[width=4.5cm]{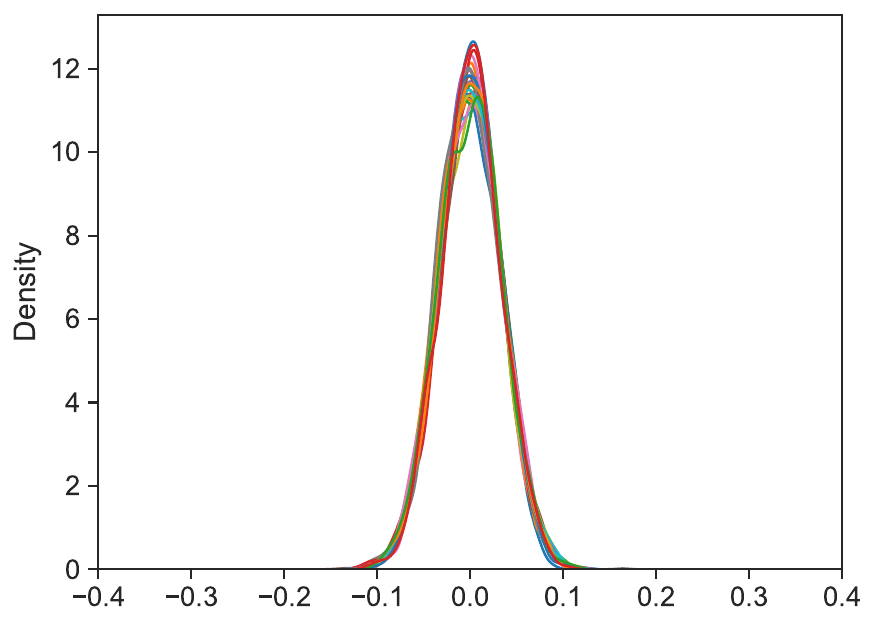}}
    \subfigure[self-attention of 13B parameters.]{
    \includegraphics[width=4.5cm]{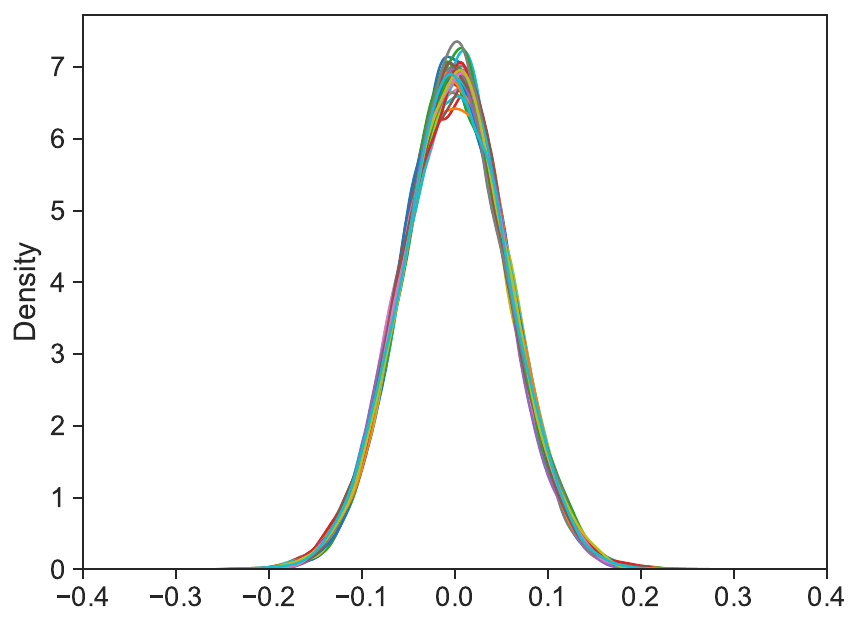}}
    \subfigure[FFN of 350M parameters.]{
    \includegraphics[width=4.5cm]{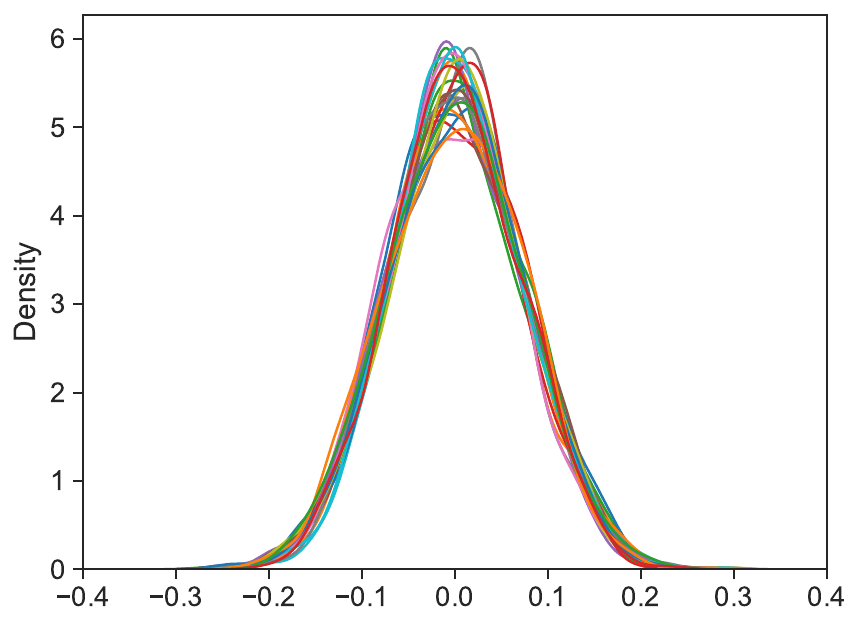}}
    \subfigure[FFN of 1.7B parameters.]{
    \includegraphics[width=4.5cm]{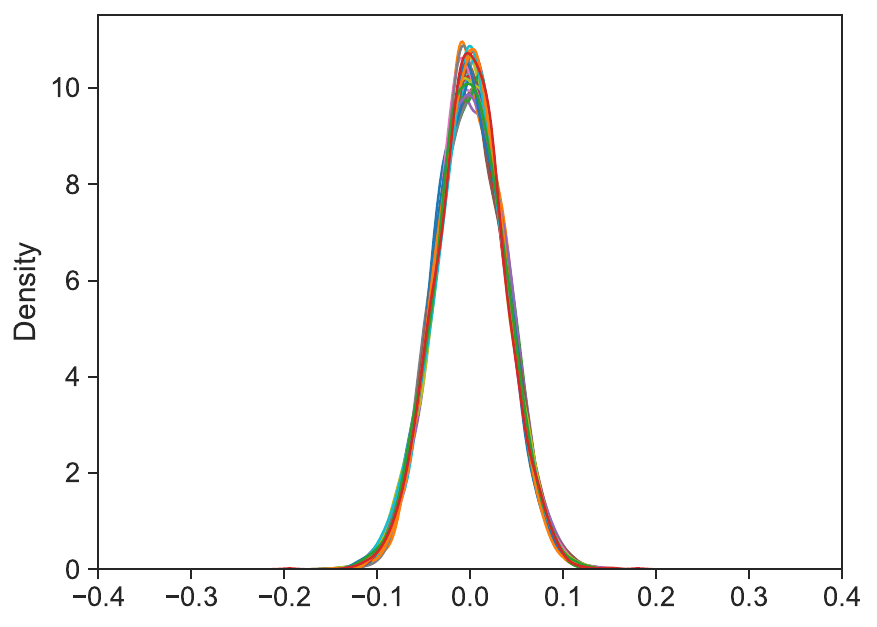}}
    \subfigure[FFN of 13B parameters.]{
    \includegraphics[width=4.5cm]{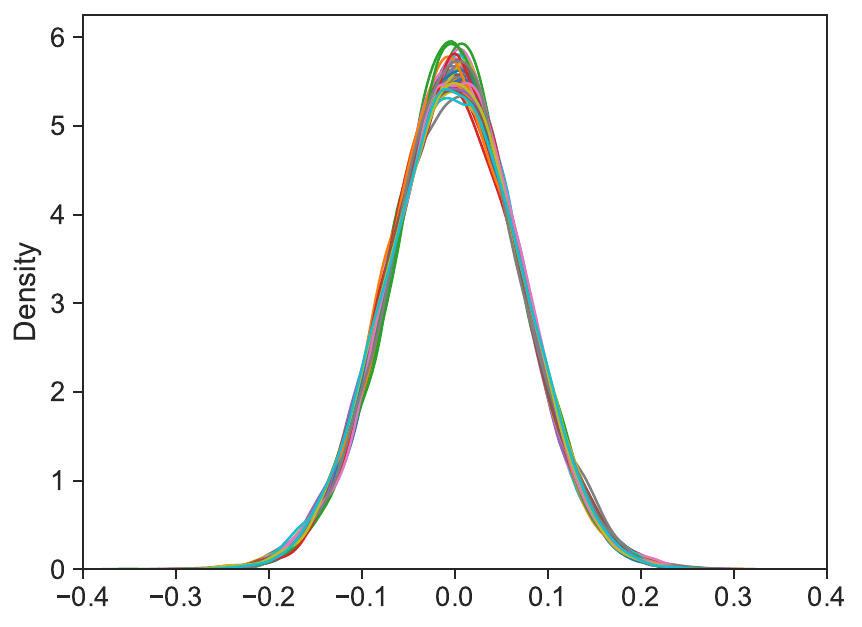}}
    \caption{Output distributions of each sub-layer at the beginning of pre-training.}
    \label{fig:output_distribution}
\end{figure}

\begin{figure}[!t]
\centering
    \subfigure[self-attention of 350M parameters.]{
    \includegraphics[width=4.5cm]{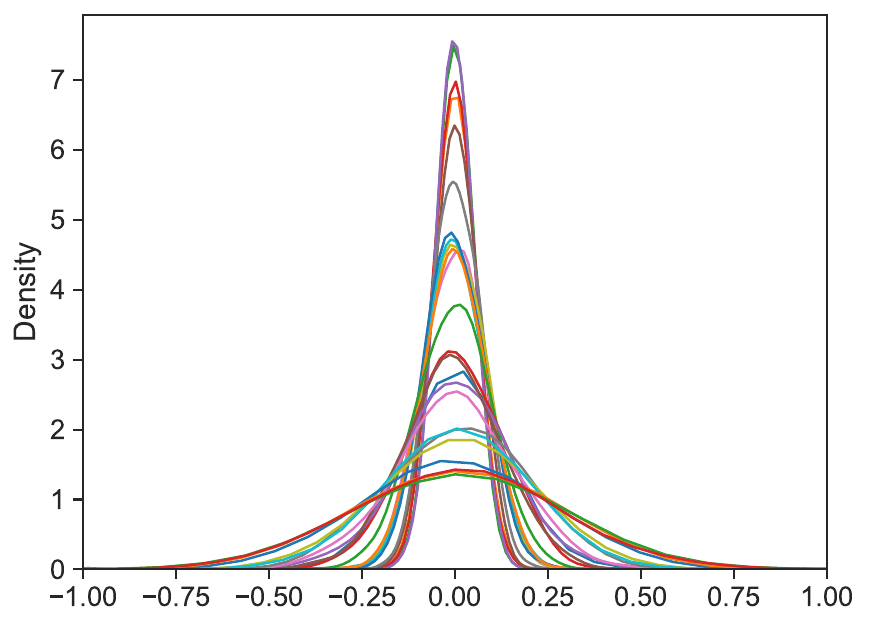}}
    \subfigure[self-attention of 1.7B parameters.]{
    \includegraphics[width=4.5cm]{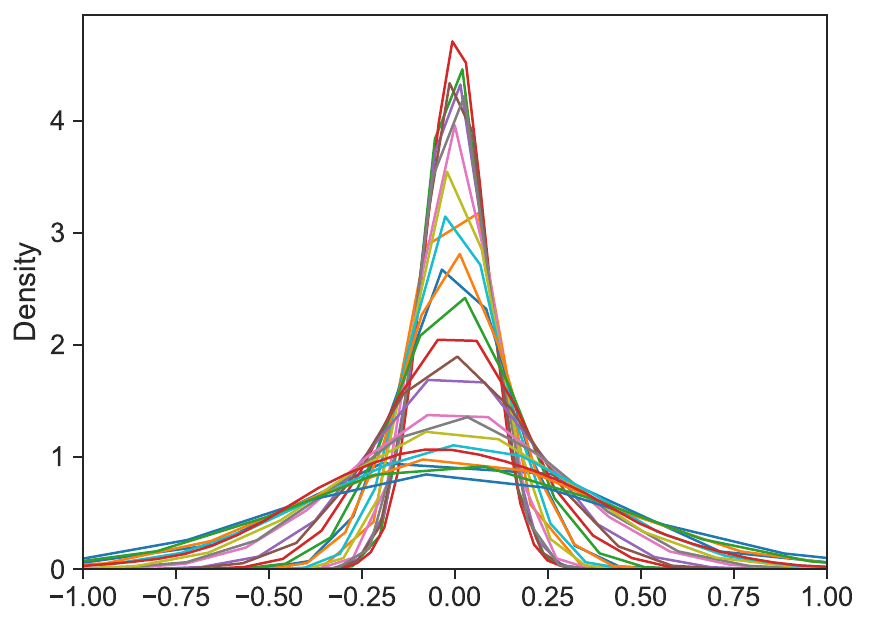}}
    \subfigure[self-attention of 13B parameters.]{
    \includegraphics[width=4.5cm]{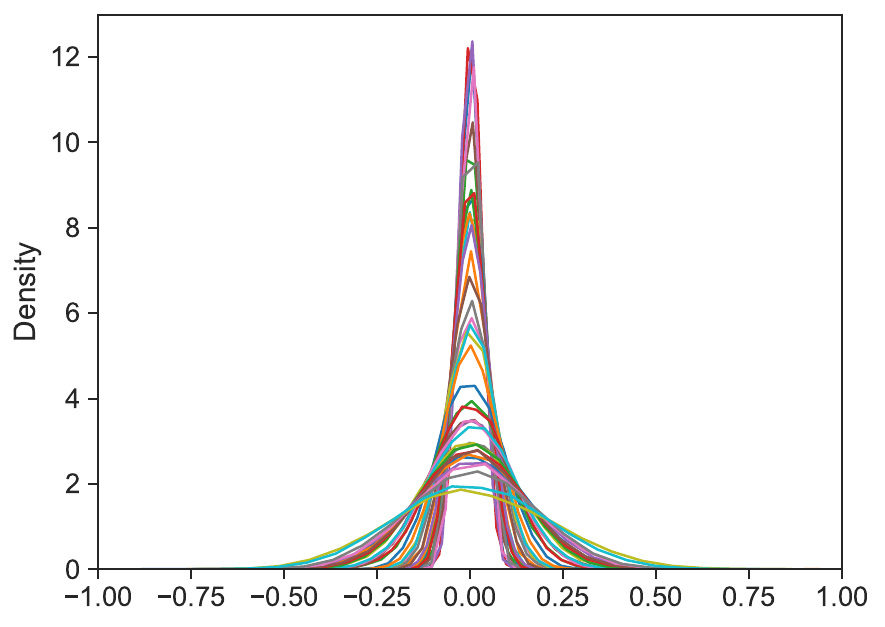}}
    \subfigure[FFN of 350M parameters.]{
    \includegraphics[width=4.5cm]{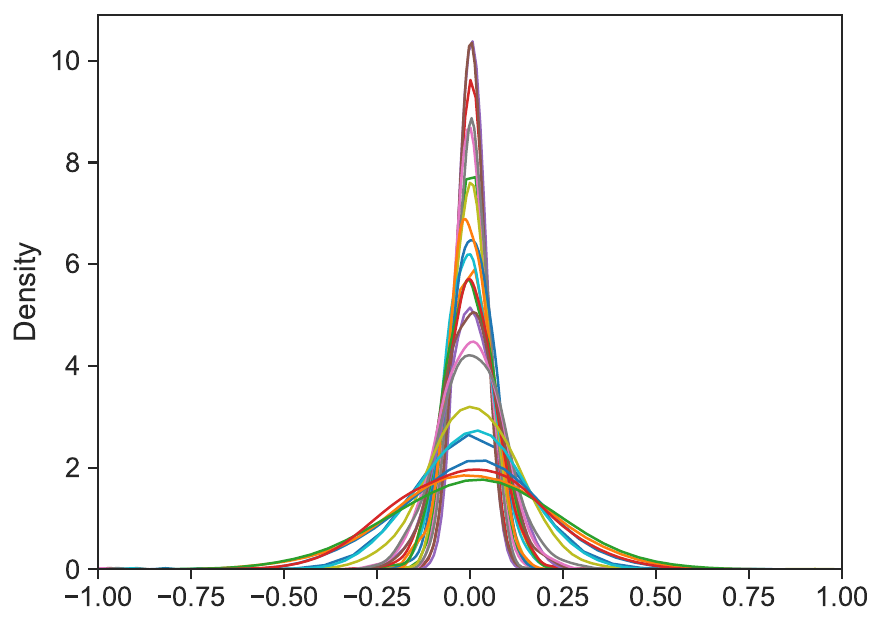}}
    \subfigure[FFN of 1.7B parameters.]{
    \includegraphics[width=4.5cm]{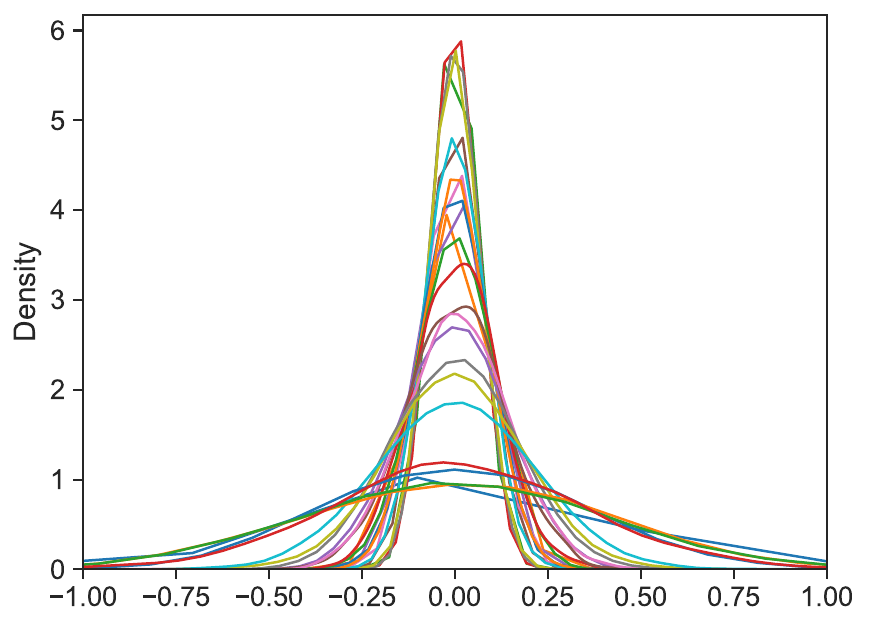}}
    \subfigure[FFN of 13B parameters.]{
    \includegraphics[width=4.5cm]{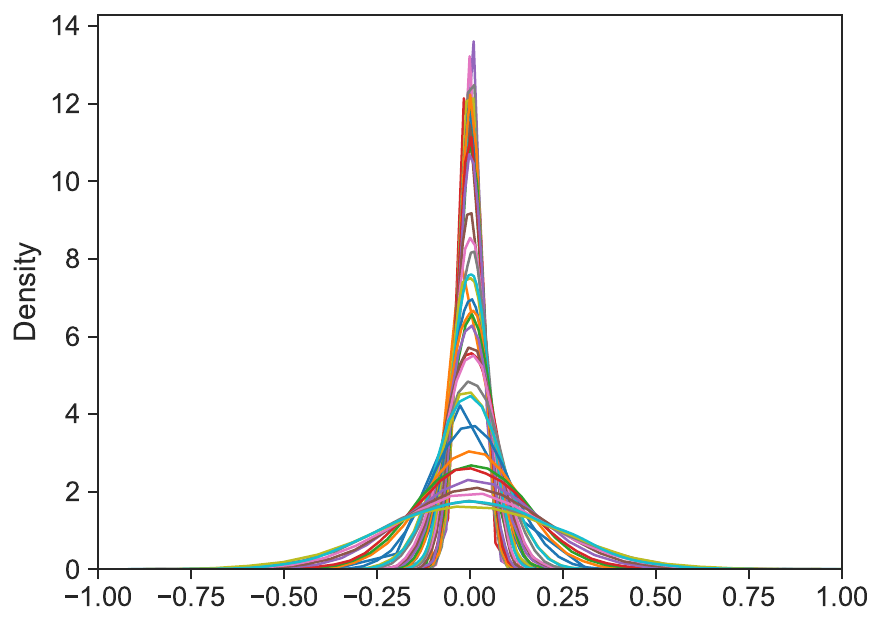}}
    \caption{Output distributions of each sub-layer after pre-training.}
    \label{fig:output_distribution_after_pretrain}
\end{figure}

Figures \ref{fig:output_distribution} and \ref{fig:output_distribution_after_pretrain} shows output distributions of each sub-layer at the beginning of pre-training and after pre-training.
These figures indicate that each sub-layer output is close to a normal distribution in various configurations both at the beginning of pre-training and after pre-training.
Therefore, the assumption in this study, which is that the vector $x$ at each layer follows the normal distribution, is reasonable.


\section{Discussion on Various Activation Functions in FFN}
\label{sec:relu_in_ffn}

\subsection{ReLU}
We consider the case where we use the ReLU function as $\mathcal{F}$ instead of the identity function.
Because we assume that parameters and the input vector at each layer follow the normal distribution, the internal layer also follows the normal distribution.
Therefore, each element of the FFN internal layer is a negative value with half probability.
In this case, we can regard that the ReLU function cuts the elements of the internal layer by half.
Thus, we replace $W_1 \in \mathbb{R}^{d_{\mathrm{ffn}} \times d}$ and $W_2 \in \mathbb{R}^{d \times d_{\mathrm{ffn}}}$ with $W_1 \in \mathbb{R}^{\frac{d_{\mathrm{ffn}}}{2} \times d}$ and $W_2 \in \mathbb{R}^{d \times \frac{d_{\mathrm{ffn}}}{2}}$ in the discussion in Section \ref{sec:analysis_on_jacobian_ffn} when we use the ReLU function as $\mathcal{F}$.

\subsection{SiLU}
We consider the case where we use the SiLU function as $\mathcal{F}$.
The definition of the SiLU function is as follows:
\begin{align}
\mathrm{SiLU}(x) = x \circ \mathrm{Sigmoid}(x)
\end{align}
where $\mathrm{Sigmoid}$ is the sigmoid function.
\begin{align}
\nonumber
\frac{\partial \mathrm{SiLU}(x)}{\partial x} &= \mathrm{Sigmoid}(x) + x \circ \frac{\partial \mathrm{Sigmoid}(x)}{\partial x} \\ \nonumber
&= \mathrm{Sigmoid}(x) + x \circ \mathrm{Sigmoid}(x) \circ (1 - \mathrm{Sigmoid}(x)) \\
& = \mathrm{Sigmoid}(x) \circ (1 + x \circ (1 - \mathrm{Sigmoid}(x)))
\end{align}
Let $D = \mathrm{diag}\Bigl(\frac{\partial \mathrm{SiLU}(W_{1}x)}{\partial W_{1}x}\Bigr) \in \mathbb{R}^{d_{\mathrm{ffn}} \times d_{\mathrm{ffn}}}$ .
Then, we obtain the Jacobian of the $\mathrm{FFN}$ as follows:
\begin{align}
\frac{ \partial \mathrm{FFN}(\mathrm{LN}(x'))}{\partial \mathrm{LN}(x')} = W_2 D W_1
\end{align}
Therefore,
\begin{align}
\bigg\|\frac{ \partial \mathrm{FFN}(\mathrm{LN}(x'))}{\partial \mathrm{LN}(x')}\bigg\|_2 \leq \|W_2\|_2 \| D \|_2 \|W_1 \|_2
\end{align}

The spectral norm of the diagonal matrix $D$ is the maximum absolute value of its diagonal elements:
\begin{align}
\| D \|_2 = \max_{i} \bigg\| \frac{\partial \text{SiLU}(x_i)}{\partial x_i} \bigg\|
\end{align}
Moreover, we find that its maximum occurs at $x \approx 2.4$ and is approximately $1.1$, and thus, $\| D \|_2 \leq 1.1$.
Because $\|W_1\|_2 \approx \sigma_1 (\sqrt{d} + \sqrt{d_{\mathrm{ffn}}})$ and $\|W_2\|_2 \approx \sigma_2 (\sqrt{d} + \sqrt{d_{\mathrm{ffn}}})$, we can express the upper bound as the following inequality:
\begin{align}
\bigg\|\frac{ \partial \mathrm{FFN}(\mathrm{LN}(x'))}{\partial \mathrm{LN}(x')}\bigg\|_2 \leq 1.1 (\sigma_1 \sigma_2 (\sqrt{d} + \sqrt{d_{\mathrm{ffn}}})^2)
\end{align}
Finally, Equation (\ref{eq:ffn_detail_upper_bound}) can be rewritten as:
\begin{align}
\bigg\|\frac{\partial y}{\partial x'}\bigg\|_2 &\leq 1 + 1.1 \biggl(\frac{\sigma_1 \sigma_2}{\sigma_{x'}} C_{\mathrm{ffn}} \biggr)
\end{align}

\subsection{SwiGLU}
We consider the case where we use the SwiGLU function as $\mathcal{F}$.
When we use the SwiGLU function, the FFN layer is expressed as follows:
\begin{align}
\mathrm{FFN}(x) = W_2 (\mathrm{Swish}(W_1 x) \circ (V x))
\end{align}
where $V \in \mathbb{R}^{d_{\mathrm{ffn}} \times d}$, and $V$ follows a normal distribution $\mathcal{N}(0, \sigma_{V}^2)$.
Then, we compute the Jacobian of the $\mathrm{FFN}(x)$ as follows:
\begin{align}
\frac{\partial \mathrm{FFN}(x)}{\partial x} = \frac{\partial \mathrm{FFN}(x)}{\partial W_1 x}\frac{\partial W_1 x}{\partial x} + \frac{\partial \mathrm{FFN}(x)}{\partial V x}\frac{\partial V x}{\partial x} \label{eq:swiglu_jacobian}
\end{align}
\begin{align}
\frac{\partial \mathrm{FFN}(x)}{\partial W_1 x} &= W_2 \bigg(\mathrm{diag}(Vx) \circ \mathrm{diag}\bigg(\frac{\partial \mathrm{Swish}(W_1 x)}{\partial W_1 x}\bigg)\bigg) \\
\frac{\partial \mathrm{FFN}(x)}{\partial V x} &= W_2 (\mathrm{diag}(\mathrm{Swish}(W_1 x)))
\end{align}
Therefore, we can rewrite Equation (\ref{eq:swiglu_jacobian}) as follows:
\begin{align}
\frac{\partial \mathrm{FFN}(x)}{\partial x} = W_2 \bigg(\mathrm{diag}(Vx) \circ \mathrm{diag}\bigg(\frac{\partial \mathrm{Swish(W_1 x)}}{\partial W_1 x}\bigg)\bigg) W_1 + W_2 (\mathrm{diag}(\mathrm{Swish}(W_1 x))) V
\end{align}
Let $J_1 = W_2 \bigg(\mathrm{diag}(Vx) \circ \mathrm{diag}\bigg(\frac{\partial \mathrm{Swish}(W_1 x)}{\partial W_1 x}\bigg)\bigg) W_1$ and $J_2 = W_2 (\mathrm{diag}(\mathrm{Swish}(W_1 x))) V$, $\frac{\partial \mathrm{FFN}(x)}{\partial x} = J_1 + J_2$.
We can derive the upper bound of $\big\| \frac{\partial \mathrm{FFN}(x)}{\partial x} \big\|_2$ as follows:
\begin{align}
\bigg\| \frac{\partial \mathrm{FFN}(x)}{\partial x} \bigg\|_2 \leq \|J_1\|_2 + \|J_2\|_2
\end{align}
For $\|J_1\|_2$, we have:
\begin{align}
\|J_1\|_2 \leq \|W_2\|_2 \|\mathrm{diag}(Vx)\|_2 \bigg\| \mathrm{diag}\bigg(\frac{\partial \mathrm{Swish}(W_1 x)}{\partial W_1 x}\bigg)\bigg\|_2 \|W_1\|_2
\end{align}
Each element of $Vx$ is a sum of $d$ independent random variables with variance $\sigma_{x}^2 \sigma_{V}^2$, and thus, $\mathrm{var}(Vx) = d\sigma_{x}^2 \sigma_{V}^2$.
Therefore, from the expected maximum of $d_{\mathrm{ffn}}$ Gaussian random variables,
\begin{align}
\|\mathrm{diag}(Vx)\|_2 \leq \sigma_x \sigma_V \sqrt{2 \ d \ \mathrm{log} \ d_{\mathrm{ffn}}}
\end{align}
The derivation of the $\mathrm{Swish}$ function is bounded by $1.1$ in the same manner as the SiLU function:
\begin{align}
\bigg\| \frac{\partial \mathrm{Swish}(W_1 x)}{\partial W_1 x}\bigg\|_2 \leq 1.1
\end{align}
The spectral norms of $W_1$ and $W_2$ can be obtained $\|W_1\|_2 \approx \sigma_1 (\sqrt{d} + \sqrt{d_{\mathrm{ffn}}})$ and $\|W_2\|_2 \approx \sigma_2 (\sqrt{d} + \sqrt{d_{\mathrm{ffn}}})$ as described in Section \ref{sec:analysis_on_jacobian_ffn}.
We can obtain the upper bound of $\|J_1\|_2$ with these equations:
\begin{align}
\|J_1\|_2 \leq 1.1 \sigma_x \sigma_V \sigma_1 \sigma_2 (\sqrt{d} + \sqrt{d_{\mathrm{ffn}}})^{2} \sqrt{2 \ d \ \mathrm{log} \ d_{\mathrm{ffn}}}
\end{align}

For $\|J_2\|_2$, we have:
\begin{align}
\|J_2\|_2 \leq \|W_2\|_2 \|(\mathrm{diag}(\mathrm{Swish}(W_1 x)))\|_2 \|V\|_2
\end{align}
Due to $|\mathrm{Swish}(W_1 x)| \leq |W_1 x|$ and $\mathrm{var}(W_1 x) = d\sigma_{x}^2\sigma_{1}^2$, we can obtain:
\begin{align}
\|(\mathrm{diag}(\mathrm{Swish}(W_1 x)))\|_2 \leq \sigma_x \sigma_1 \sqrt{2 \ d \ \mathrm{log} \ d_{\mathrm{ffn}}}
\end{align}
Therefore,
\begin{align}
\|J_2\|_2 \leq \sigma_x \sigma_V \sigma_1 \sigma_2 (\sqrt{d} + \sqrt{d_{\mathrm{ffn}}})^{2} \sqrt{2 \ d \ \mathrm{log} \ d_{\mathrm{ffn}}}
\end{align}
Based on these equations, we can derive the upper bound as follows:
\begin{align}
\bigg\|\frac{\partial \mathrm{FFN}(x)}{\partial x}\bigg\|_2 &\leq \|J_1\|_2 + \|J_2\|_2 \nonumber \\
& = 2.1 \ \sigma_x \sigma_V \sigma_1 \sigma_2 (\sqrt{d} + \sqrt{d_{\mathrm{ffn}}})^{2} \sqrt{2 \ d \ \mathrm{log} \ d_{\mathrm{ffn}}} \nonumber \\
& = \sigma_x \sigma_V \sigma_1 \sigma_2 C_{\mathrm{swiglu}} \label{eq:swiglu_upper_bound}
\end{align}
where $C_{\mathrm{swiglu}}$ includes $2.1 \ (\sqrt{d} + \sqrt{d_{\mathrm{ffn}}})^{2} \sqrt{2 \ d \ \mathrm{log} \ d_{\mathrm{ffn}}}$ for the simplification.

Finally, we consider the actual Transformer layer that includes layer normalization and residual connection:
\begin{align}
\bigg\|\frac{\partial y}{\partial x'}\bigg\|_2 &\leq 1 + \frac{\sigma_V \sigma_1 \sigma_2}{\sigma_{x'}} C_{\mathrm{swiglu}}
\end{align}
We note that $\sigma_x$ in Equation (\ref{eq:swiglu_upper_bound}) is equal to $1$ in the actual Transformer layer because we apply the layer normalization to the input of the FFN layer.

\section{Relation between Input Length and the Standard Deviation of Self-Attention}
\label{sec:std_selfattn}

% especially at the early stage of the pre-training. と言うかは謎
We explain that the standard deviation of the self-attention layer becomes small as the input length is long.
Because we assume that parameters and the input vector at each layer follow the normal distribution, the expectation of each element of $(W_{Qi} \ x)^\mathrm{T} (W_{Ki} \ X)$ is $0$.
Therefore, the expectation after the softmax function is $\frac{1}{L}$ where $L$ is the length of the input sequence.
Thus, the long input sequence decreases the standard deviation of the self-attention layer.

To simplify, we consider the case where the number of self-attention heads is $1$.
In this case, we can obtain the variance of each calculation with the following equation.
\begin{align}
\mathrm{var}(W_{O} (x)) &= \mathrm{var}(W_O) \mathrm{var}(x) \ d \\
\mathrm{var}(W_{V} (x)) &= \mathrm{var}(\mathrm{W_V}) \mathrm{var}(x) \ d
\end{align}
where $\mathrm{var}$ represents the variance of the matrix/vector.
Thus, the variance of the self-attention layer, $\mathrm{var}(\mathrm{Attn}(x))$, is as follows:
\begin{align}
\mathrm{var}(\mathrm{Attn}(x)) &= \mathrm{var}(W_O) \ d \sum^{L} \frac{\mathrm{var}(\mathrm{W_V}) \mathrm{var}(x) \ d}{L^2} \\
&= \frac{\mathrm{var}(W_O) \mathrm{var}(\mathrm{W_V}) \mathrm{var}(x)d^2}{L}
\end{align}

\section{Details on Jacobian Matrix of Self-Attention}
\label{sec:detail_selfattn_jacobian}
We can represent $\mathrm{concat}(\mathrm{head}_1(x), ..., \mathrm{head}_h(x))$ with the summation of the matrix multiplications as follows:
\begin{align}
\mathrm{concat}(\mathrm{head}_1(x), ..., \mathrm{head}_h(x)) = \sum_{i=1}^h \mathrm{head}_i W_i
\end{align}
where $W_i \in \mathbb{R}^{d_{\mathrm{head}} \times d}$ whose corresponding element is $1$ and the others are $0$.
Let $J^{i}$ be the Jacobian of the $\mathrm{head}_i(x)$, we can represent $J^Z$ in Section \ref{sec:analysis_on_jacobian_attn} as follows:
\begin{align}
J^Z = \sum_{i=1}^h J^i W_i 
\end{align}
In addition, the self-attention consists of the interaction among inputs and outputs of each position in the sequence.
Thus, we add indices to Jacobians to represent the positions of the input and output.
Let $x_j$ be the input of the position $j$, and $z_{k}^{i}$ be the $i$-th head of the output position $k$, and $J_{kj}^{i} = \frac{\partial z_{k}^{i}}{\partial x_j}$.
Because $J^Z$ can be regarded as the Jacobian of the input of the position $j$, we can convert Equation (\ref{eq:tmp_head_jacobian}) into the following Equation:
\begin{align}
J^Z = \sum_{k=1}^{L} \sum_{i=1}^h J_{kj}^i W_i \label{eq:tmp_head_jacobian}
\end{align}
where $L$ is the length of the input and output sequences.
Therefore, we compute $J_{kj}^i$ to obtain $J^Z$ in Section \ref{sec:analysis_on_jacobian_attn}.

We can obtain a head of the output position $k$, i.e., $z_{k}$, as follows\footnote{To simplify the equations, we omit the index $i$ to represent $i$-th head from the head and parameters.}
\begin{align}
z_{k} = \sum_{l=1}^{L}A_{kl}v_{l}
\end{align}
where $A_{kl}$ is the $l$-th element of the attention vector, $\mathrm{softmax}\left(\frac{(W_Q\ x_k)^{\mathrm{T}} (W_K \ X)}{\sqrt{d_{\mathrm{head}}}}\right)$ and $v_{l}$ is $W_{V} x_{l}$.
Therefore, to obtain $J_{kj}$, we differentiate $z_{k}$ with respect to $x_j$ as:
\begin{align}
J_{kj} = \frac{\partial z_{k}}{\partial x_{j}} = \sum_{l=1}^{L}\left( \frac{\partial A_{kl}}{\partial x_j}{v_{l}}^{\mathrm{T}} + A_{kl}\frac{\partial v_l}{\partial x_j} \right)
\end{align}
\begin{align}
\frac{\partial v_l}{\partial x_j} &= W_V \delta_{lj} \\
\delta_{lj} &=
\begin{cases}
1 \ \  \text{if $l = j$} \\
0 \ \ \text{otherwise}
\end{cases}
\end{align}
Thus,
\begin{align}
\frac{\partial z_{k}}{\partial x_{j}} = \sum_{l=1}^{L}\left( \frac{\partial A_{kl}}{\partial x_j}{v_{l}}^{\mathrm{T}}\right) + A_{kj}W_V \label{eq:jacobian_before_assumption}
\end{align}

Here, we assume that the attention vector is uniform.
In this assumption, since $A_{kj} = \frac{1}{L}$, we can obtain the spectral norm of the second term for Equation (\ref{eq:jacobian_before_assumption}) as $\| A_{kj} W_V \|_2 \approx \frac{\sigma_V}{L}(\sqrt{d} + \sqrt{d_{\mathrm{head}}})$, where $\sigma_V$ is the standard deviation of $W_V$.
To calculate the first term, we compute $\frac{\partial A_{kl}}{\partial x_j}$.
\begin{align}
\frac{\partial A_{kl}}{\partial x_j} &= A_{kl}\left(\frac{\partial S_{kl}}{\partial x_j} - \sum_{m=1}^{L}A_{km}\frac{\partial S_{km}}{\partial x_j} \right) \\
\frac{\partial S_{kl}}{\partial x_j} &= \frac{1}{\sqrt{d_{\mathrm{head}}}}\left(W_{Q}^{\mathrm{T}} W_{K}x_l \delta_{kj} + W_{K}^{\mathrm{T}} W_{Q} x_k \delta_{lj} \right)
\end{align}
Let $D_l$ be $W_{Q}^{\mathrm{T}}W_{K}x_l$ and $E_k$ be $W_{K}^{\mathrm{T}} W_{Q} x_k$. Then,
\begin{align}
\frac{\partial S_{kl}}{\partial x_j} &= \frac{1}{\sqrt{d_{\mathrm{head}}}} \left(D_l \delta_{kj} + E_k \delta_{lj} \right)
\end{align}
Therefore,
\begin{align}
\frac{\partial A_{kl}}{\partial x_j} &= \frac{A_{kl}}{\sqrt{d_{\mathrm{head}}}}\left((D_l \delta_{kj} + E_k \delta_{lj}) - \sum_{m=1}^{L} A_{km} (D_m \delta_{kj} + E_k \delta_{mj}) \right) \\
&= \frac{A_{kl}}{\sqrt{d_{\mathrm{head}}}} \left(\delta_{lj}E_k - A_{kj} E_{k} + \delta_{kj}\left(D_l - \sum_{m=1}^{L}A_{km}D_m \right) \right) \label{eq:unfolded_attention_derivative}
\end{align}
We assign Equation (\ref{eq:unfolded_attention_derivative}) to the first term of Equation (\ref{eq:jacobian_before_assumption}) and use the assumption $A_{kj} = \frac{1}{L}$:
\begin{align}
\sum_{l=1}^{L}\left( \frac{\partial A_{kl}}{\partial x_j}{v_{l}}^{\mathrm{T}}\right) &= \sum_{l=1}^{L}\left(\frac{A_{kl}}{\sqrt{d_{\mathrm{head}}}} \left(\delta_{lj}E_k - A_{kj} E_{k} + \delta_{kj}\left(D_l - \sum_{m=1}^{L}A_{km}D_m \right) {v_{l}}^{\mathrm{T}} \right) \right) \\
&= \frac{1}{L\sqrt{d_{\mathrm{head}}}}\sum_{l=1}^{L}\left(\delta_{lj}E_k - \frac{1}{L} E_{k} + \delta_{kj}\left(D_l - \sum_{m=1}^{L}\frac{1}{L}D_m \right) {v_{l}}^{\mathrm{T}} \right) \\
&= \frac{1}{L\sqrt{d_{\mathrm{head}}}}\left( \sum_{l=1}^{L}\left( \left(\delta_{lj}E_k + \delta_{kj}D_l \right) {v_{l}}^{\mathrm{T}} \right) + \left(\sum_{l=1}^{L}\left(-\frac{1}{L}E_{k} - \sum_{m=1}^{L}\frac{1}{L}D_m \right) {v_{l}}^{\mathrm{T}} \right) \right) \\
&= \frac{1}{L \sqrt{d_{\mathrm{head}}}} \left(E_{k} {v_{j}}^{\mathrm{T}} + \sum_{l=1}^{L}\left(\delta_{kj}D_l {v_{l}}^{\mathrm{T}} \right) - \left(E_{k} + \sum_{m=1}^{L}D_{m} \right){v_{l}}^{\mathrm{T}} \right) \\
&= \frac{1}{L \sqrt{d_{\mathrm{head}}}} \left(E_{k} {v_{j}}^{\mathrm{T}} - E_{k}{v_{l}}^{\mathrm{T}} + \delta_{kj}\sum_{l=1}^{L}\left(D_{l}{v_{l}}^{\mathrm{T}}\right) - \left(\sum_{m=1}^{L}D_{m} \right){v_{l}}^{\mathrm{T}} \right)
\end{align}
Based on the assumption that parameters and the vector at each layer follow the normal distribution, we assume that the mean of $D_{l}$, $E_{k}$, and $v_{l}$ is $0$, and thus, we can obtain their norms from their variances.
In addition, we assume that the standard deviation of $W_Q$, $W_K$ and $W_V$ is $\sigma$.
Then, $\mathrm{var}(D_l) = \mathrm{var}(E_k) = d d_{\mathrm{head}} \sigma^{4}$, $\mathrm{var}(\sum_{l}^{L} D_l) = Ldd_{\mathrm{head}}\sigma^{4}$, and $\mathrm{var}(v_l) = d \sigma^{2}$.
Thus, $\|E_{k}v_{j}^{\mathrm{T}} \|_2 \approx \| E_{k}v_{l}^{\mathrm{T}} \|_2 \approx \sigma^{3}\sqrt{d^{3}d_{\mathrm{head}}^{2}}$, $\|\sum_{l=1}^{L} (D_{l}v_{l}^{\mathrm{T}}) \|_2 \approx \|(\sum_{m=1}^{L} D_m) v_{l}^{\mathrm{T}} \|_2 \approx \sigma^3\sqrt{Ld^3d_{\mathrm{head}}^2}$.
Therefore,
\begin{align}
& \Bigg\|\sum_{l=1}^{L}\left( \frac{\partial A_{kl}}{\partial x_j}{v_{l}}^{\mathrm{T}}\right)\Bigg\|_2 \nonumber \\
& \leq \frac{1}{L \sqrt{d_{\mathrm{head}}}} \left(\Big\| E_{k} {v_{j}}^{\mathrm{T}} \Big\|_2 + \Big\| E_{k}{v_{l}}^{\mathrm{T}} \Big\|_2 + \delta_{kj}\Bigg\|\sum_{l=1}^{L}\left(D_{l}{v_{l}}^{\mathrm{T}}\right) \Bigg\|_2 + \Bigg\| \left(\sum_{m=1}^{L}D_{m} \right){v_{l}}^{\mathrm{T}} \Bigg\|_2 \right) \\
& \approx \frac{1}{L\sqrt{d_{\mathrm{head}}}}\left(2\sigma^{3}\sqrt{d^{3}d_{\mathrm{head}}^{2}} + (\delta_{kj} + 1)\sigma^3\sqrt{Ld^3d_{\mathrm{head}}^2} \right) \\
&= \frac{1}{\sqrt{L}}\left(\delta_{kj} + 1 + \frac{2}{\sqrt{L}}\right)\sigma^3\sqrt{d^3d_{\mathrm{head}}}
\end{align}
Based on this Equation, we can compute the upper bound of $\|J_{kj}\|_2$:
\begin{align}
\|J_{kj}\|_2 &= \Bigg\| \sum_{l=1}^{L}\left( \frac{\partial A_{kl}}{\partial x_j}{v_{l}}^{\mathrm{T}}\right) + A_{kj}W_V \Bigg\|_2 \nonumber \\
& \leq \Bigg\| \sum_{l=1}^{L}\left( \frac{\partial A_{kl}}{\partial x_j}{v_{l}}^{\mathrm{T}}\right) \Bigg\|_2 + \| A_{kj}W_V \|_2 \\
& \leq \frac{1}{\sqrt{L}}\left(\delta_{kj} + 1 + \frac{2}{\sqrt{L}}\right)\sigma^3\sqrt{d^3d_{\mathrm{head}}} + \frac{\sigma}{L}(\sqrt{d} + \sqrt{d_{\mathrm{head}}})
\end{align}
Moreover, we can compute the upper bound of $\|J^Z\|_2$ from Equation (\ref{eq:tmp_head_jacobian}) as follows:
\begin{align}
\|J^Z\|_2 &= \bigg\| \sum_{k=1}^{L} \sum_{i=1}^h J_{kj}^i W_i \bigg\|_2 \\
& \leq \sum_{k=1}^{L} \sum_{i=1}^h \| J_{kj}^i\|_2 \|W_i\|_2 \\
& \approx h \left(\frac{1}{\sqrt{L}}\left(1 + L + \frac{2L}{\sqrt{L}}\right)\sigma^3\sqrt{d^3d_{\mathrm{head}}} + \frac{L\sigma}{L}(\sqrt{d} + \sqrt{d_{\mathrm{head}}})\right) \\
& = h \left(\left(\sqrt{L} + 2 + \frac{1}{\sqrt{L}} \right)\sigma^3\sqrt{d^3d_{\mathrm{head}}} + \sigma(\sqrt{d} + \sqrt{d_{\mathrm{head}}}) \right)
\end{align}

\section{Comparison with Post-LN Transformer}
\label{sec:post-ln}

As described in Section \ref{sec:preln}, recent studies use the Pre-LN Transformer architecture to construct their LLMs because the architecture is more stable.
In contrast, some recent studies reported that the Post-LN Transformer, which is the original architecture, can achieve better performance than the Pre-LN if we address the instability issue in the Post-LN, i.e., the vanishing gradient problem~\citep{liu-etal-2020-understanding,takase-etal-2023-b2t,https://doi.org/10.48550/arxiv.2203.00555}.
We discuss whether the Pre-LN Transformer entirely underperforms the Post-LN.
We conducted experiments on machine translation experiments because previous studies mainly focused on them.

We followed the experimental settings in \citet{takase-etal-2023-b2t}.
Table \ref{tab:mt_hyper_params} shows the details of hyper-parameters.
We used the WMT English-to-German training dataset~\citep{NIPS2017_7181,ott-etal-2018-scaling}, and evaluated each model in newstest2010-2016.
We used the encoder-decoder architecture proposed by \citet{cross_selfattn_encdec}.
To stabilize the Post-LN Transformer, we applied DeepNet~\citep{https://doi.org/10.48550/arxiv.2203.00555} and B2T connection~\citep{takase-etal-2023-b2t}.
We compared them to Scaled Embed, that is, the Pre-LN Transformer with the stabilizing techniques described in this paper.

Table \ref{tab:bleu_comp_postln} shows the averaged BLEU scores among newstest2010-2016.
For the BLEU score calculation, we used SacreBLEU~\citep{post-2018-call} to obtain compatible scores~\citep{marie-etal-2021-scientific}.
The signature of SacreBLEU is \texttt{BLEU+case.mixed+numrefs.1+smooth.exp+tok.13a+version.1.5.0}.
As shown in this table, we used two learning rates: $\mathrm{lr}=1.0 \times 10^{-3}$ and $3.0 \times 10^{-3}$.
For $\mathrm{lr}= 1.0 \times 10^{-3}$, DeepNet and B2T connection outperformed Scaled Embed.
Thus, the Post-LN Transformer-based methods achieved better performance than the Pre-LN Transformer-based method.
This result corresponds to reports in previous studies~\citep{liu-etal-2020-understanding,https://doi.org/10.48550/arxiv.2203.00555,takase-etal-2023-b2t}.

On the other hand, for $\mathrm{lr}= 3.0 \times 10^{-3}$, Scaled Embed achieved better performance than the others with $\mathrm{lr}=1.0 \times 10^{-3}$, and the training of the others failed due to the exploding gradients.
This result indicates that the Pre-LN Transformer-based method can achieve better performance if we use a large learning rate.
Therefore, the Pre-LN Transformer (with the stabilizing techniques) is more stable than the Post-LN Transformer-based method, and thus, it can achieve better performance when we use a large learning rate that is too large to train the Post-LN Transformers.

\begin{table}[!t]
  \centering
  \caption{Hyper-parameters used in the comparison with Post-LN Transformer.}
  \label{tab:mt_hyper_params}
  \begin{tabular}{ l | c } \hline
  Name & Value \\ \hline
  Layer num & 18 \\
  Hidden dim size & 512 \\
  FFN dim size & 2048 \\
  Attention heads & 8 \\
  Dropout rate & 0.5 \\
  Precision & \texttt{float16} \\
  Word dropout rate & 0.1 \\
  Max tokens & 7168 \\
  Adam $\beta_1$ & 0.9 \\
  Adam $\beta_2$ & 0.98 \\
  Gradient clipping & 0.1 \\
  $lr$ decay style & inverse square root \\
  Warmup step & 4000 \\
  Weight decay & 0 \\ \hline
  \end{tabular}
\end{table}

\begin{table}[!t]
  \centering
  \footnotesize
  \caption{Averaged BLEU scores among newstest2010-2016.}
  \label{tab:bleu_comp_postln}
  \begin{tabular}{ l | c c c c c c c | c  } \hline
  Model & 2010 & 2011 & 2012 & 2013 & 2014 & 2015 & 2016 & Average $\uparrow$\\ \hline \hline
  \multicolumn{9}{c}{$\mathrm{lr}=1.0 \times 10^{-3}$} \\ \hline \hline
  DeepNet & \textbf{24.65} & 22.30 & \textbf{22.87} & 26.51 & 27.29 & 29.77 & 34.87 & 26.89 \\
  B2T connection & 24.46  & \textbf{22.42} & 22.85 & 26.51 & \textbf{27.46} & 29.91 & 34.65 & 26.89 \\ 
  Scaled Embed & 24.32 & 22.21 & 22.40 & 26.38 & 26.89 & \textbf{29.98} & 34.53 & 26.67 \\ \hline \hline
  \multicolumn{9}{c}{$\mathrm{lr}=3.0 \times 10^{-3}$} \\ \hline \hline
  DeepNet & \multicolumn{7}{c|}{N/A} & N/A \\
  B2T connection & \multicolumn{7}{c|}{N/A} & N/A \\ 
  Scaled Embed & 24.52 & 22.23 & 22.86 & \textbf{26.54} & 27.35 & 29.90 & \textbf{35.16} & \textbf{26.94} \\ \hline  
  \end{tabular}
\end{table}

\end{document}